\documentclass[lettersize,journal]{IEEEtran}
\usepackage{amsmath,amsfonts}
\usepackage{algorithmic}

\usepackage{algorithm}
\usepackage{array}
\usepackage[caption=false,font=normalsize,labelfont=sf,textfont=sf]{subfig}
\usepackage{textcomp}
\usepackage{booktabs}
\usepackage{makecell} 
\usepackage{multirow}
\usepackage{csquotes} 
\usepackage{stfloats}
\usepackage{utfsym}
\usepackage{url}
\usepackage{verbatim}
\usepackage{graphicx}
\graphicspath{{./fig/}}
\usepackage{url}
\usepackage{cite}

\makeatother
\usepackage{hyperref} 
\usepackage{algorithm}
\usepackage{amsmath}
\usepackage{amssymb}
\usepackage{algorithmic}
\usepackage{tikz}
\usepackage{pifont}
\hypersetup{
colorlinks=true,
linkcolor=red,
citecolor=green,
urlcolor = black
} 
\begin{document}

\title{A Point-Neighborhood Learning Framework for Nasal Endoscopic Image Segmentation}

\author{Pengyu Jie, Wanquan Liu,~\IEEEmembership{Senior Member,~IEEE,}, Chenqiang Gao$^*$, Yihui Wen, Rui He, Weiping Wen, \\Pengcheng Li, Jintao Zhang, Deyu Meng,~\IEEEmembership{Senior Member,~IEEE,}
        \thanks{Pengyu Jie, Wanquan Liu, Chenqiang Gao$^*$ and Jintao Zhang are with the School of Intelligent Engineering, Sun Yat-Sen University-Shenzhen Campus, Shenzhen 518107, China (e-mail: jiepy3@mail2.sysu.edu.cn; gaochq6@mail.sysu.edu.cn). Yihui Wen, Rui He and Weiping Wen are with Department of Otolaryngology, The First Affiliated Hospital of Sun Yat-sen University, Guangzhou 510000, China. Pengcheng Li is with Chongqing University of Posts and Telecommunications, Chongqing 400065, China. Deyu Meng is with the School of Mathematics and Statistics, Xi’an Jiaotong University, Xi’an, 710049, China. ($^*$Corresponding author)}} 

\markboth{Journal of \LaTeX\ Class Files,~Vol.~14, No.~8, August~2021}%
{Pengyu Jie \MakeLowercase{\textit{et al.}}: A Point-Neighborhood Learning Framework for Nasal Endoscopic Image Segmentation}

 
\maketitle 

\begin{abstract}
    Lesion segmentation on nasal endoscopic images is challenging due to its complex lesion features.
    Fully-supervised deep learning methods achieve promising performance with pixel-level annotations but impose a significant annotation burden on experts.
    Although weakly supervised or semi-supervised methods can reduce the labelling burden, their performance is still limited.
    Some weakly semi-supervised methods employ a novel annotation strategy that labels weak single-point annotations for the entire training set while providing pixel-level annotations for a small subset of the data.
    However, the relevant weakly semi-supervised methods only mine the limited information of the point itself, while ignoring its label property and surrounding reliable information.
    This paper proposes a simple yet efficient weakly semi-supervised method called the Point-Neighborhood Learning (PNL) framework.
    PNL incorporates the surrounding area of the point, referred to as the point-neighborhood, into the learning process.
    In PNL, we propose a point-neighborhood supervision loss and a pseudo-label scoring mechanism to explicitly guide the model’s training.
    Meanwhile, we proposed a more reliable data augmentation scheme.
    The proposed method significantly improves performance without increasing the parameters of the segmentation neural network.
    Extensive experiments on the NPC-LES dataset demonstrate that PNL outperforms existing methods by a significant margin. Additional validation on colonoscopic polyp segmentation datasets confirms the generalizability of the proposed PNL.

\end{abstract}

 
\section{Introduction}
\label{sec:introduction}
\IEEEPARstart{N}{asopharyngeal} carcinoma (NPC) is a common and hard-to-treat malignancy in the head and neck, and the incidence of NPC has seen an upward trend, partly attributed to improved screening methods and better awareness in clinical practice\cite{NPC_review_TransBionologyReview}.
\begin{figure}[t]
	\centering
	\includegraphics[width=\columnwidth]{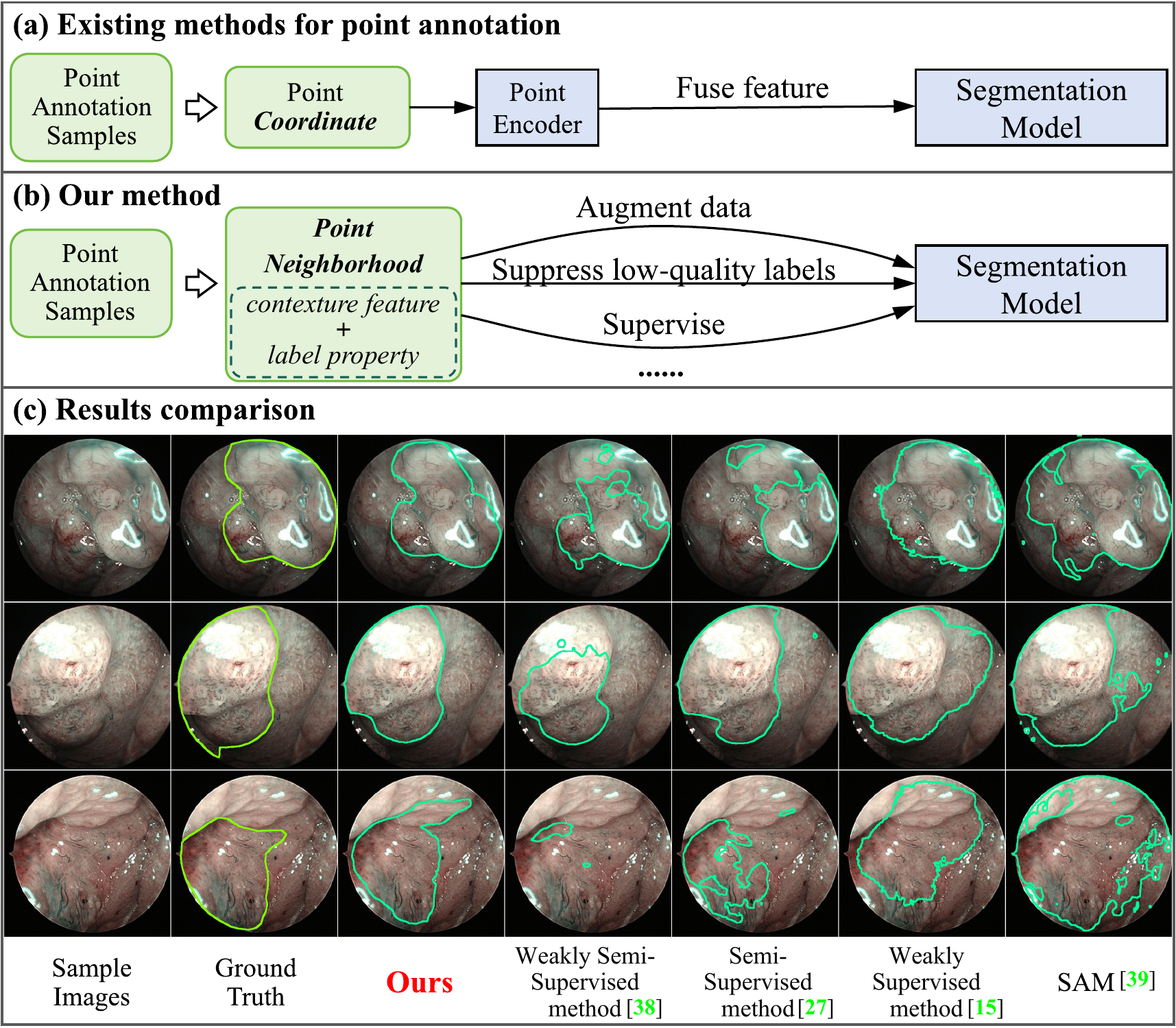}
	\caption{Comparison between the existing image segmentation methods (a) and our proposed method (b) for point annotation. Methods (a) use point coordinates as input, while our method (b) mines the points' neighborhoods for data augmentation, supervision, and low-quality pseudo-label suppression.
    (c) shows a visual comparison of the test results for different methods.
    }
	\label{fig:introduction_figure}  
    \vspace{-0.55cm}
\end{figure}
Currently, the NPC lesion localization on endoscopic images is still widely through manual screening, which heavily relies on the expertise and experience of experts. Thanks to deep learning, automatic lesion segmentation can help to locate potential NPC lesion areas quickly. Applying segmentation techniques to endoscopic images has attracted widespread attention, and many deep-learning methods have been developed\cite{NPC_MOHAMMED_4,NPC_MOHAMMED_5,NPC_NiXiaoguang_1,NPC_NiXiaoguang_2,NPC_Li2018}.
While these methods have shown promising performance, they require a substantial amount of high-quality pixel-level annotations for fully-supervised learning which is time-consuming and labor-intensive. 
Compared to common segmentation tasks, annotating NPC lesion areas is much more challenging\cite{ganeshan2024enhancing}.
The NPC lesions are visually similar to non-lesions\cite{Xu2024}, with random shapes, diverse textures and uncertain boundaries (see samples in Fig.~\ref{fig:introduction_figure}(c)), which require annotators to reach the level of NPC diagnostic experts to guarantee annotation quality.


Weakly supervised learning is a widely used strategy to alleviate the annotation burden which labels all training samples with weak annotations, e.g., image-level\cite{weak_CAM,weak_SEAM}, scribble curve\cite{scribble_ZhuangXH,scribble_transform}, bounding-box\cite{bbox_BCM,Cheng_2023_CVPR} and point\cite{Yu_2022_CVPR,point_OD_point_teaching,point_Group_RCNN,Wu_2024_CVPR,LI2021107979}.
As the weak labels can not be directly used to train segmentation models, the commonly adopted strategy is to design mining methods to automatically covert the weak labels into pixel-level labels, known as pseudo-labels\cite{Yu_2022_CVPR,weak_SEAM,7775087,10597372,Fan2023,LI2021107979}.
Besides the pseudo-label methods, the quality of pseudo-labels is also heavily affected by the difficulty of the segmentation tasks\cite{Cheng_2023_CVPR}.
When the task involves distinguishing similar foreground and background regions., e.g., the NPC lesion segmentation discussed in this paper\cite{Xu2024}, the pseudo-labels may be of low quality, which would significantly hinder model training, and lead to low performance\cite{Cheng_2023_CVPR}.


Semi-supervised learning\cite{semi_SEMI_survey,semi_GTA_TS,semi_double_tch_14,semi_double_tch_21,semi_mixup_BCP,semi_error_localization,semi_CANet,semi_temporal} is another widely-used strategy to alleviate the annotation burden, which only annotates a part of training samples with pixel-level labels, keeping the rest unlabeled. 
To training a segmentation model, a commonly adopted strategy is to firstly train a initial segmentation model with pixel-level annotated samples, and then apply it to mine the pixel-level labels of the unlabeled samples, also known as pseudo-labels. Finally, the mined pseudo-labels are used further to retrain the model\cite{semi_mixup_BCP,semi_pseudolabel,semi_MCF,semi_SEMI_survey}.
Similar to the weakly supervised learning, the quality of the pseudo-labels is heavily affected by the difficulty of the segmentation task. As mentioned previously, the NPS lesion segmentation is a relatively difficult task, which could lead to low quality of pseudo-labels, and negatively affect the model training. Additionally, the distribution shift between the labeled samples and the unlabeled samples also negatively affects the model training, as discussed in\cite{9941371,Liu_2022_CVPR}, so it is often to assume that they have the same distribution, which does not always holds in practice.

The weakly semi-supervised learning can be considered as a combination of the weakly supervised and semi-supervised learning through adding weak labels to all the training samples of semi-supervised learning\cite{Wu_2024_CVPR,point_OD_point_DETR,point_OD_point_DETR_3DOD,point_WangHong_MICCAI,point_WangHong_MIA}, which is an effective method to improve the performance of the segmentation model without heavily adding annotation burden. Recently, Point-SEGTR\cite{point_WangHong_MIA} adopted this learning setting with point annotation for NPC lesion segmentation, and obtained better performance than both weakly supervised and semi-supervised learning methods.
However, similar to the point prompt of the popular Segment Anything Model (SAM)\cite{point_SAM}, the spatial coordinates of the point annotation are directly fed into a point encoder to learn the feature (Fig.~\ref{fig:introduction_figure}(a)), which ignores its label property and reliable information surrounding the point. 
Actually, although it contains the label property, the point annotation is very weak, which introduces many challenges for NPC lesion segmentation. For example, the point annotation can hardly contain texture and color information\cite{PointSupervision}, both of which are very important for NPC lesion segmentation. Additionally, the point annotation is so weak that it is hard to be used as supervised signal for model training\cite{Cheng_2022_CVPR}. Furthermore, some widely validated effective strategies, e.g., sample mixup\cite{other_mixup}, cannot be used to the point annotation samples directly.

To address above challenges, in this paper, we propose an explicit yet effective Point-Neighborhood Learning (PNL) method based on teacher-student framework to train any existing supervised segmentation model for the NPC lesion segmentation with the weakly semi-supervised configuration. 
We argue that the small neighborhood of the point annotation, e.g., a circular area centered at the point, is much likely the part of the ground truth. Intuitively, this is reasonable because the NPC lesion area usually is relatively large, and annotators are prone to annotate the points at the interior area of the ground truth\cite{PointSupervision,Yu_2022_CVPR}. 
In practice, this assumption can hold just through simply training annotators who usually have the expertise level.
Thus, the neighborhood of the point annotation fully falls into the ground truth with high confidence. 
In this way, we can confidently transform the single-point annotation into a neighborhood annotation without any annotation burden. 
This transformation can greatly reduce the gap between pixel-level and point annotation samples, which makes the pseudo-label mining relatively easier. Meanwhile, it can make the point annotation samples to be more effectively learned (Fig.~\ref{fig:introduction_figure}(b)). Specifically, 
(1) we utilize point-neighborhoods as supervision signals, focusing on predictions within these areas while ignoring external pixels. 
(2) Point-neighborhoods are used as a powerful constraint to suppress pseudo-labels of low quality which are far away from the annotated points. 
(3) The data augmentation are applied with the point-neighborhoods. We build a point-neighborhood bank to store high confident positive samples based on the shape consistency of point-neighborhoods and use it to mixup data to extend data diversity. Finally, we adopt teacher-student\cite{semi_pseudolabel} framework to optimize the target model. Our contributions are mainly three-folds:
\begin{itemize}
    \item To efficiently learn on weakly semi-supervision datasets, we propose an explicit but effective method: Point-Neighborhood Learning (PNL). We transform point annotations to point-neighborhoods and then explicitly exploit the lesion feature information in the point-neighborhoods.
    \item Based on point-neighborhood transformation, we propose Point-Neighborhood Supervision (PNS) and Pseudo-label Scoring Mechanism (PSM) to provide more reliable supervision. To expand the data feature's diversity, we build Point-Neighborhood Mixup (PNMxp) and Particular-Value Randomly Mixup (PVRMxp) strategies to augment data. 
    \item Comprehensive experiments on nasal endoscopic datasets demonstrate that our method outperforms state-of-the-art (SOTA) approaches, while validation on three colonoscopic polyp datasets showcases its generalization ability and effectiveness in small object segmentation.
\end{itemize}


\section{related work} 
\label{sec_relatedwork}
Medical image segmentation, particularly for complex tasks like NPC lesion segmentation, has advanced significantly through deep learning. To alleviate the annotation burden, various weakly supervised, semi-supervised, and weakly semi-supervised strategies have been explored. This section provides a brief overview of these approaches.
\begin{figure*}[ht]
	\centering 
    \vspace{-0.5cm}
	\includegraphics[height=5.6cm]{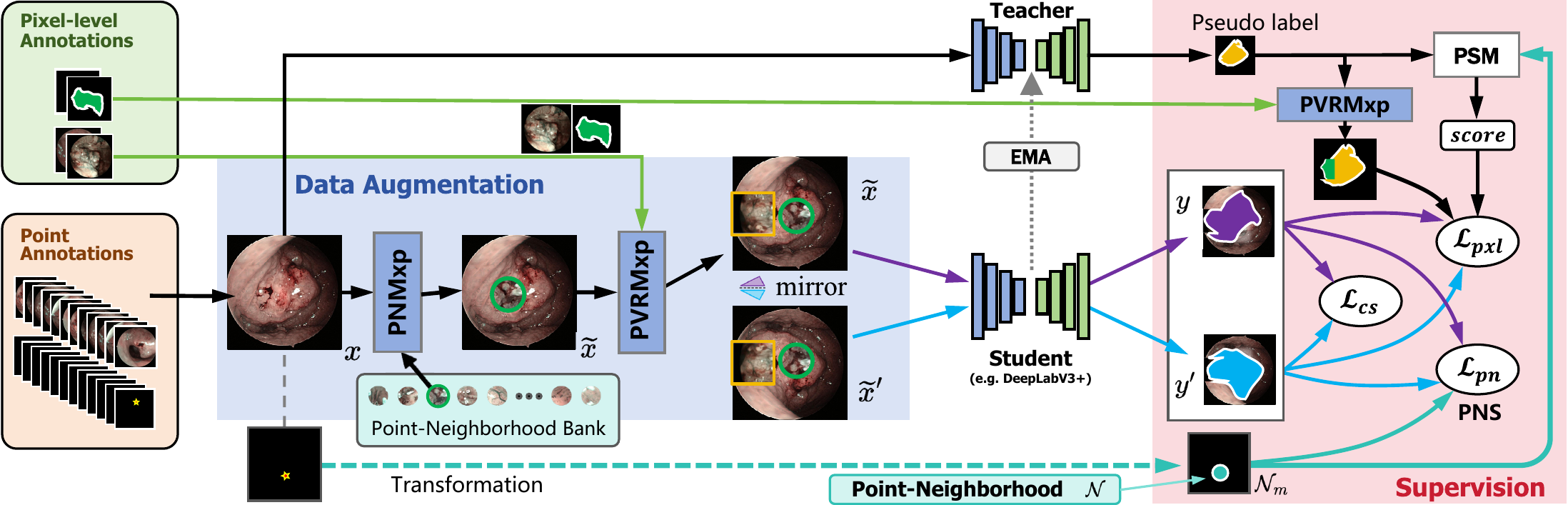} 
	\caption{Framework of Our Method. Leveraging point-neighborhood transformations, we propose strategies for two critical training stages: data augmentation and supervision. In the data augmentation, Point-Neighborhood Mixup (PNMxp) and Particular-Value Randomly Mixup (PVRMxp) enhance data diversity, then improving the model generalization. In the supervision, Point-Neighborhood Supervision (PNS) provides local supervision for point-neighborhood regions. Additionally, the Pseudo-Label Scoring Mechanism (PSM) scores and suppresses low-confidence pseudo-labels.}
	\label{fig:overall_framework}
    \vspace{-0.45cm} 
\end{figure*}
\subsection{Weakly Supervised learning}
Weakly supervised learning is a commonly approach to reduce the annotation burden by providing weak labels for training samples, e.g., image-level, scribbles, boxes and points.
To obtain full supervision labels, existing methods routinely exploit the weak annotations to mine pixel-level pseudo-labels.
Image-level annotation only annotates the objects' classes.
For image-level annotations, some methods\cite{weak_ADELE,Chen_2022_CVPR,Chen2_2022_CVPR} generated pseudo-labels using the regions with strong responses to classification inferring.
Scribble annotation annotates samples' location with hand-drawn curves. 
The representative methods include conditional random fields methods\cite{wang2019boundary} and pseudo-label-based self-supervised methods\cite{scribble_transform,scribble_RGB_D1,scribble_ZhuangXH}.
Box annotation annotates the object's position with rectangular boxes.
Recently, some methods\cite{bbox_BCM,Cheng_2023_CVPR} proposed promising box-to-mask strategies to mine available pseudo-labels for supervision.
Point annotation refers to labeling a specific point (single-pixel) inside the object in an image.
Yu et al.\cite{Yu_2022_CVPR} proposed a point refinement method to mine new pseudo-point-labels to work as subsequent supervision basis. 
Ying et al.\cite{point_supervision_infrared} proposed to evolved single-point labels into more accurate pseudo-labels through iterative learning of the model. 
\cite{point_supervision_SOD_floodfill,point_supervision_SOD_floodfill_2} use flood filling to expand points.
Of the four types of weak annotations, point annotation has a lower annotation burden while retaining crucial location and label information.
However, current point annotation methods often treat points as isolated signals, ignoring the rich contextual information in their surrounding regions which is well explored in this paper.

\subsection{Semi-supervised learning}

Semi-supervised learning uses a small amount of labeled data and a larger set of unlabeled data, often relying on pseudo-label generation and iterative refinement to maximize the use of unlabeled data. Tarvainen et al.\cite{semi_EMA} introduced the Mean Teacher framework, where the teacher is trained on labeled data and generates pseudo-labels for the student. The student, in turn, refines the teacher’s knowledge through Exponential Moving Average (EMA)\cite{semi_EMA,semi_mixup_BCP,semi_error_localization,semi_GTA_TS} updates.
Bai et al.\cite{semi_mixup_BCP} developed a copy-paste processing technique to transfer labeled data features to unlabeled samples, generating reliable pseudo-labels. Typically, pseudo-labels do not fully align with the ground truth, then hinder training. To address this issue, Kwon et al.\cite{semi_error_localization} explored potential structural relationships in unlabeled data to improve pseudo-labels. Similarly, Jin et al.\cite{semi_GTA_TS} proposed a strategy for pseudo-label generation and filtering, then reduce low-quality pseudo-labels.
Pseudo-label effectiveness varies by task. In complex tasks like NPC lesion segmentation, low-quality pseudo-labels can hinder model training. We explore a method based on point-neighborhoods to mitigate the impact of low-quality pseudo-labels.

\subsection{Weakly Semi-supervised learning with point annotation}
Weakly semi-supervised learning combines weakly and semi-supervised approaches to balance annotation cost and segmentation accuracy.
There exist some attempts\cite{10120949,point_supervision_devil} to tackle the weakly semi-supervised task using point annotations.
Zhang et al.\cite{point_Group_RCNN} localized the object through a grouping strategy that extends the points into complete representations of the object region and then continuously optimizes the pseudo-labels.
Some methods\cite{point_OD_point_DETR,point_OD_point_DETR_3DOD,point_WangHong_MICCAI} used point-encoder to encode point coordinates, and then query feature maps for inference. 
Recently, Point-SEGTR\cite{point_WangHong_MIA} was proposed to segment nasal endoscopic images inspired by point-based weakly semi-supervised methods\cite{point_OD_point_DETR,point_WangHong_MICCAI}. Point-SEGTR trained the point-querying model on pixel-level samples and generated pseudo-labels for point-level samples for pseudo-supervision.
However, these methods only used the location information of the point annotation itself, while ignoring its label property.

In this paper, we transform the point annotations to point-neighborhood annotations which effectively support multiple learning strategies on the point-annotated samples, including the supervision learning, data augmentation, and training with high-quality pseudo-labels, which make our method obviously outperforms SOTA methods.

\section{Method}
\label{sec_framework}

\subsection{Overview}
Our method is illustrated in Fig.~\ref{fig:overall_framework}, which can be used to train existing segmentation models, e.g., DeepLabV3+\cite{backbone_DeepLabV3Plus}, PSPNet\cite{backbone_PSPNet}, SegNet\cite{backbone_Segnet} and SegFormer\cite{backbone_segformer}. Assuming that the training set is $D$ = \{$D_1$, $D_2$\}, where $D_1$ and $D_2$ are the subset of pixel-level and point-level annotated samples. In our notation, $(x_{D_1},y_{D_1}),(x_{D_2},y_{D_2})$ represent the samples from the dataset $D_1,D_2$, respectively. Following the mainstream methods\cite{semi_mixup_BCP,semi_double_tch_21,semi_MCF,Liu_2022_CVPR}, we adopt the Mean Teacher framework\cite{semi_EMA} to iteratively mine pseudo-labels and to train models.
Specifically, the teacher and student are set as the same segmentation model and initiated with random parameters respectively.
The teacher model firstly learns on $D_1$ and mines pseudo-labels for $D_2$. Then we use both $D_1$ and $D_2$ to train the student model, and further update the teacher via the EMA\cite{semi_EMA}:
\begin{equation} \label{eq:EMA}
    \setlength{\abovedisplayskip}{3pt} 
    \setlength{\belowdisplayskip}{3pt}
    \resizebox{0.45\hsize}{!}{$\begin{aligned}
    \Theta_{t}^{t} = \alpha \Theta_{t-1}^{t} + \left( 1- \alpha\right) \Theta^{s}_{t},
\end{aligned}$}
\end{equation}
where $t$ represents the $t$-th time step, and $ \alpha $ indicates the decaying factor. The $\Theta^t$ and $\Theta^s$ mean the weights of the teacher and student models, respectively.

As detailedly discussed in Section~\ref{sec:introduction}, the surrounding pixels in the point-neighborhood belong to the same object as the point-annotated pixel with high probability\cite{PointSupervision,Yu_2022_CVPR}.
Thus, we transform the single point to a circle point-neighborhood $\mathcal{N}$ with the radius $\mathcal{R} $ to compensate for the sparsity of point annotations.
We denote the point-neighborhood label matrix as $\mathcal{N}_m$ in which we set the pixels inside $\mathcal{N}$ as 1 while others as 0 (one-hot coding).
Base on this point-neighborhood transformation, we can design effective learning strategies in this paper. 
First, we propose two kinds of data augmentation strategies, namely, Point-Neighborhood Mixup (PNMxp) and Particular-Value Randomly Mixup (PVRMxp), and augment samples.
Second, based on the point-neighborhood annotations, we propose a Pseudo-label Scoring Mechanism (PSM) to suppress the low-quality pseudo-labels. 
Third, as the point-neighborhood annotations contain much more supervision signals than single-point annotations, we leverage them to directly supervise model's training. 

\subsection{Point-Neighborhood Supervision (PNS)}\label{subsec:PNT}

\begin{figure}[h]
    \centering
    \includegraphics[width=8.01cm]{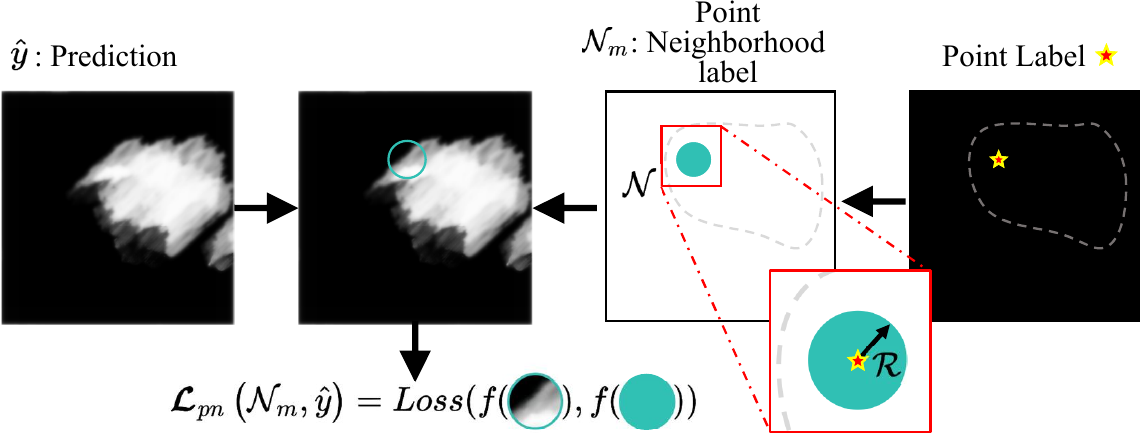} 
    \caption{The illustration of PNS loss. PNS loss measures the prediction accuracy inside $\mathcal{N} $ but ignores those pixels outside of $\mathcal{N} $.}
    \label{fig:PNS}
\end{figure}

The point annotation can work as a powerful supervisor, indicating that the pixel at the point belongs to the object\cite{Laradji_2018_ECCV,PointSupervision}.
However, supervising solely with a single pixel is insufficient\cite{point_supervision_infrared} because it lacks contextual information. 
The point-neighborhood transformation introduces local contextual information, e.g., textures, colors, to learning process which can supervise the model's training.
We propose Point-Neighborhood Supervision (PNS, illustrated in Fig.~\ref{fig:PNS}) to supervise inside $\mathcal{N}$.
We approximate the pixels within the point-neighborhood by setting their labels to 1 (foreground) to facilitate the construction of the loss function.
We execute PNS supervision with L1 loss:
\begin{equation}
    \setlength{\abovedisplayskip}{3pt} 
    \setlength{\belowdisplayskip}{3pt}
    \resizebox{0.76\hsize}{!}{$\begin{aligned}
    \mathcal{L}_{pn}(\hat{y},\mathcal{N}_m)=\frac{1}{\pi \mathcal{R}^2} \sum_{(i,j)\in \mathcal{N}} \left|\mathcal{N}_m(i,j)-\hat{y}(i,j)\right|,
\end{aligned}$}
\end{equation}
where $\hat{y}$ indicates the prediction of the network model.

Point-neighborhood transformations may be contaminated by non-lesion regions, leading to noise interference in the model. We employ L1 loss to assess point-neighborhood supervision due to its robustness against noise and outliers. In NPC lesion segmentation, the lesion boundaries are often irregular and visually similar to surrounding normal tissues, making mislabeling of non-lesion areas during training more problematic. L1 loss effectively mitigates the impact of such errors, enhancing the model's robustness. It provides smoother gradients, preventing excessive adjustments to large errors, and facilitates the rapid learning of key features of the lesion regions, especially when point annotations are sparse. This enables the model to better capture localized features of the lesions, particularly small and ambiguous ones. 
PNS is introduced to supervise the classification of pixels in the point-neighborhoods and prevent foreground pixels (lesion) from being misclassified as background pixels (non-lesion), thereby reducing the occurrence of false negative (FN) prediction.

\subsection{Dual Mixup Strategy (DM)}\label{subsec:DualMixup} 
The Mixup strategy\cite{other_mixup} generates new image-label pairs through a linear combination of two samples: $\tilde{x}=\sigma \cdot x_0+(1-\sigma) \cdot x_1$, $\tilde{y}=\sigma \cdot y_0+(1-\sigma) \cdot y_1$ where the $\sigma$ samples from the beta distribution and $(x_0, y_0)$ and $(x_1, y_1)$ are image-label pairs. 
By interpolating between the samples, Mixup increases the training data diversity. This approach enhances the model's generalization ability and improves robustness against overfitting\cite{pmlrverma19a}. However, directly mixup-ing can lead to label ambiguity, where some pixels in the mixed label $\tilde{y}$ may be assigned to multiple categories\cite{Guo_Mao_Zhang_2019,pmlrverma19a}.
To address this issue and further improve model's generalization ability, we proposed dual mixup (DM), including PNMxp and PVRMxp.
 
    \subsubsection{Point-Neighborhood Mixup (PNMxp)}\label{subsub:PNMxp}
    
\begin{figure}[ht]
        \centering
        \includegraphics[width=\columnwidth]{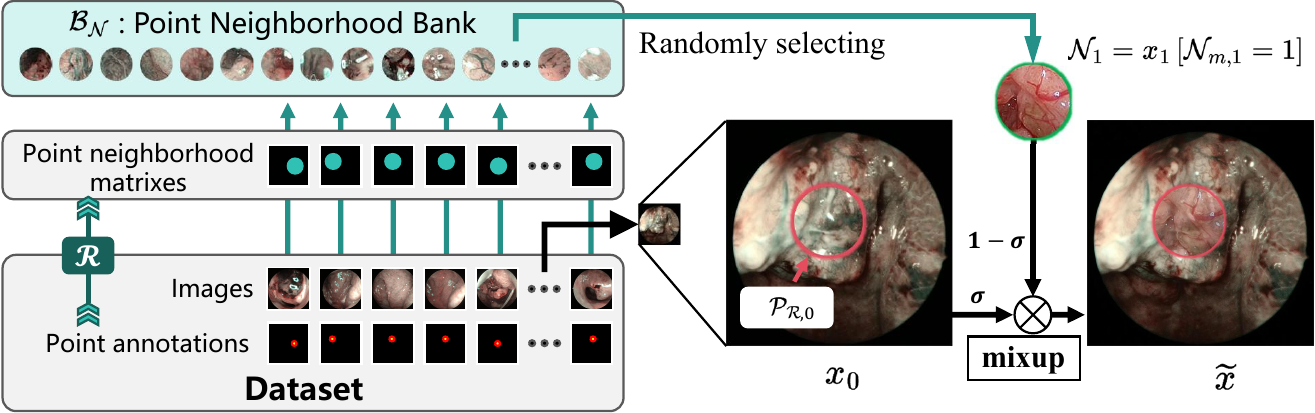}
        \caption{The illustration of Point-Neighborhood Mixup (PNMxp).}
        \label{fig:PNMxp}
    \end{figure}

    To mitigate label ambiguity, some existing methods crop the foreground along its boundary and combine it with another sample\cite{10378660}. However, this approach necessitates pixel-level annotations for all samples, which is often impractical in weakly semi-supervised tasks. Thanks to the point-neighborhood transformation, each sample of the training set has a reliable pixel-level label with a highly consistent shape. In this basis, we propose Point-Neighborhood Mixup (PNMxp) to circumvent ambiguous labels as illustrated in Fig.~\ref{fig:PNMxp}. Because the point-neighborhoods are consistently circular, there is no risk of label ambiguity when PNS is supervising pixels within these regions.
    \begin{equation}
        \resizebox{0.9\hsize}{!}{$\begin{aligned}
            &\widetilde{x}_0\left[  \mathcal{N}_{m,0}=0 \right] =  x_0\left[  \mathcal{N}_{m,0}=0 \right], \\
            &\widetilde{x}_0\left[  \mathcal{N}_{m,0}=1 \right] = \sigma \cdot x_0 \left[  \mathcal{N}_{m,0}=1 \right]+(1-\sigma) \cdot x_1 \left[ \mathcal{N}_{m,1}=1 \right],\\
            &\widetilde{y}_0 = y_0.
    \end{aligned}$}
    \end{equation}
    where the symbol function, $\left[\cdot \right]$, refers to the elements those satisfy the condition ($\cdot$). 

    In PNMxp, only the pixels within the point-neighborhoods are linearly interpolated with another sample, while those outside the point-neighborhood are ignored. To facilitate this process, we constructed a point-neighborhood bank, $\mathcal{B_{\mathcal{N}}}$, to store all the point-neighborhood slices from the images. 
    Especially when processing a large number of training samples, it provides a mechanism for centralized storage and rapid access to point-neighborhood data, avoiding the repeated calculation of the same neighborhood information, thereby saving computing resources and time.
 
    \subsubsection{Particular-Value Randomly Mixup (PVRMxp)}
\begin{figure}[ht]
        \centering
        \includegraphics[width=\columnwidth]{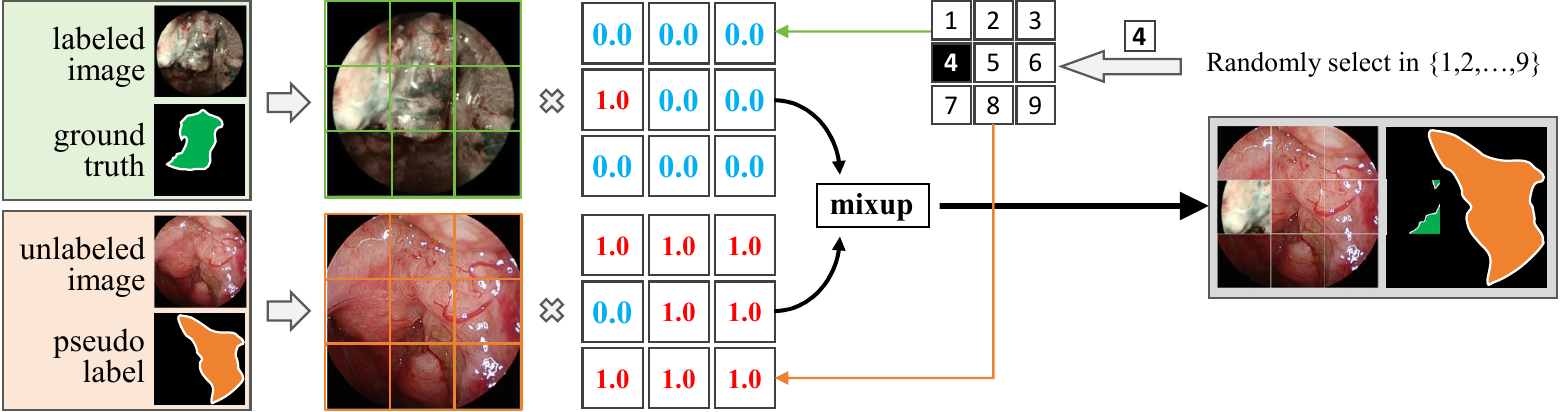}
        \caption{The illustration of Particular-Value Randomly Mixup (PVRMxp).} 
        \label{fig:PVRMxp}
    \end{figure}

    Pixel-level annotated samples $(x_{D_1}, y_{D_1})$, though a small proportion of the dataset, provide crucial supervisory information\cite{Chen_2021_CVPR, semi_mixup_BCP}, highlighting texture features and boundary patterns between lesions and non-lesions.
    Training solely on $D_2$ causes the model to converge to local optima\cite{bmvc_DASD}, often predicting all pixels as foreground, with this tendency becoming more pronounced as the proportion of $D_1$ decreases.
    We mixup $(x_{D_1},y_{D_1})$ and $(x_{D_2},y_{D_2})$ to ensure each training sample contains reliable boundary features from $x_{D_1}$. To avoid the uncertain pixels mentioned in Section~\ref{subsec:DualMixup}, we set $\sigma$ to particular values $\{0.0,1.0\}$.
    We partition samples $(x_{D_1},y_{D_1})$ and $(x_{D_2},y_{D_2})$ into nine-grid patches, randomly selecting one patch from $(x_{D_1},y_{D_1})$ and combining it with the remaining patches from $(x_{D_2},y_{D_2})$ using PVRMxp for augmentation.
    We propose the augmentation strategy Particular-Value Randomly Mixup (PVRMxp) (illustrated in Fig.~\ref{fig:PVRMxp}), where the model learns both the pixel-level annotations from $D_1$ and pseudo-annotations from $D_2$. This encourages the model to refine pseudo-labels using knowledge from $(x_{D_1}, y_{D_1})$. After PVRMxp, each augmented sample contains a random portion of $x_{D_1}$ and the pixel-level pseudo-label from $x_{D_2}$. PVRMxp expands the data space and alleviates boundary perception issues due to the limited scale of $D_1$.
    
\subsection{Pseudo-label Scoring Mechanism (PSM)}\label{subsec:PSM}
The Mean Teacher framework iteratively generates pseudo-labels using the teacher model and the quality of the pseudo labels is very important for training. Low-quality pseudo-labels can mislead the student model, degrading its performance. To mitigate this issue, we propose the Pseudo-label Scoring Mechanism (PSM), which effectively suppresses low-quality pseudo-labels, preventing them from negatively impacting the student model's learning process.
\begin{figure}[ht]
    \centering
    \includegraphics[width=8cm]{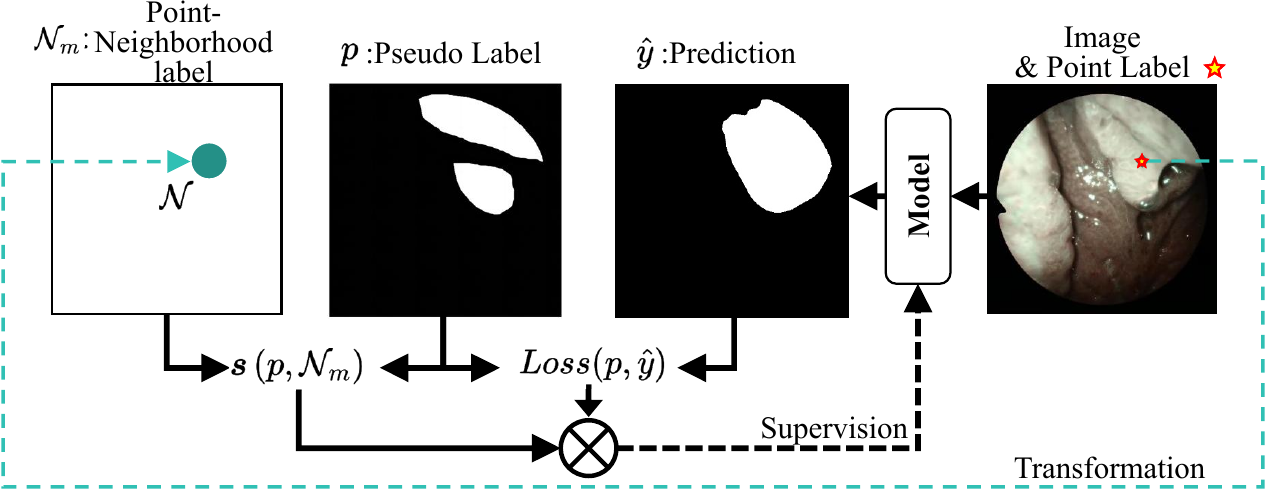}
    \caption{The proposed PSM. PSM scores pseudo-labels based on point-neighborhood and weights the pseudo-labels supervision loss with the score.}
    \label{fig:PSM}
\end{figure}
As illustrated in Fig.~\ref{fig:PSM}, the Pseudo-label Scoring Mechanism (PSM) leverages the strong prior knowledge of point-neighborhoods to assess the quality of the pseudo-labels. Pseudo-labels that align well with the point-neighborhoods are assigned high scores, while those that poorly match are assigned low scores.
The score $\boldsymbol{s}$ belongs to the interval $\left[ 0,1 \right]$.
$  \boldsymbol{s}$ is calculated as follows:
\begin{equation}
    \setlength{\abovedisplayskip}{3pt} 
    \setlength{\belowdisplayskip}{3pt}
    \boldsymbol{s}(p,\mathcal{N}_m) = \frac{\sum_{(i,j)}p \left[ p \left[ \mathcal{N}_{m,(i,j)}=1 \right] >0.5 \right]}{\sum_{(i,j)} \mathcal{N}_{m,(i,j)} },
\end{equation}
where $p $ denotes the pseudo-label. 
The pseudo-labeled samples which are assigned low scores are restrained in training until next update. 
These pseudo-labels are still used for pseudo-supervision, but their loss $ \mathcal{L}_{pxl} $ is suppressed by being weighted with $ \boldsymbol{s}$.
In the training process, PSM and PNS are independent of each other. When the pseudo-label of a sample is suppressed, the PNS will not be affected. 

\subsection{Loss Function}\label{subsec:PNS}
Our method incorporates three kinds of supervised loss functions to ensure accurate lesion region segmentation. These include the pixel-level supervision loss $ \mathcal{L}_{pxl} $, which is applied to the model's overall prediction, the point-neighborhood loss $ \mathcal{L}_{pn} $, and the consistency loss $ \mathcal{L}_{cs}$ between symmetrical outputs. During training, the pixel-level supervision loss is defined as the binary cross-entropy loss, as follows:
\begin{equation} \label{eq:LossFunc_pseudo_supervision}
    \setlength{\abovedisplayskip}{3pt} 
    \setlength{\belowdisplayskip}{3pt}
    \resizebox{0.81\hsize}{!}{$\begin{aligned}
    \mathcal{L}_{pxl}(y,\hat{y})=-\frac{1}{HW}  \left[\hat{y} \log(y) + (1-\hat{y}) \log(1-y)\right].
\end{aligned}$}
\end{equation}
where $H,W$ represent the height and width of the sample images.
The symmetrical consistency loss between outputs ($\hat{y}$ and $\hat{y}^{\prime}$) enforces consistency across symmetrical transformations of inputs ($x$ and $x^{\prime}$), encouraging stable and reliable predictions. The loss is computed using mean square error (MSE), defined as follows:
\begin{equation} \label{eq:LossFunc_sym_cs}
    \setlength{\abovedisplayskip}{3pt} 
    \setlength{\belowdisplayskip}{3pt}
    \resizebox{0.48\hsize}{!}{$\begin{aligned}
    \mathcal{L}_{cs}(\hat{y}, \hat{y}^{\prime})=-\frac{1}{H W}  \left(\hat{y}- \hat{y}^{\prime} \right)^2,
    \end{aligned}$}
\end{equation}
The supervision of the teacher model is denoted as:
\begin{equation}\label{eq:Loss_LT}
    \setlength{\abovedisplayskip}{5pt} 
    \setlength{\belowdisplayskip}{5pt}
    \resizebox{0.88\hsize}{!}{$\begin{aligned}
        \mathcal{L}_t &=  \lambda \cdot \left(
        \mathcal{L}_{pxl}\left(\Theta^t\left(x_{D_1} \right),y_{D_1}\right)+ \mathcal{L}_{pxl}\left(\Theta^t\left({x_{D_1}}^{\prime}\right),{y_{D_1}}^{\prime}\right)\right)
        \\
        + &(1-\lambda)\cdot  \mathcal{L}_{cs}\left(\Theta^t\left(x_{D_1} \right)^{\prime},\Theta^t\left({x_{D_1}}^{\prime}\right)\right)\\
        + & (1-\lambda) \cdot \left(\mathcal{L}_{pn}\left(\Theta^t\left(x_{D_1} \right),\mathcal{N}_ {m}\right)
        + \mathcal{L}_{pn}\left(\Theta^t\left({x_{D_1}}^{\prime}\right),\mathcal{N}^{\prime}_ {m}\right)\right),
    \end{aligned}$}
\end{equation}
where $x^{\prime} $ denotes the mirror of $x$. All the samples are preprocessed by PNMxp. We define $\lambda$ to indicate the weighted hyperparameter that adjusts different supervision losses.

The student model is trained on $ D_2$. $ \mathcal{L}_{pxl} $ is weighted by PSM scores, $\boldsymbol{s}(p,\mathcal{N}_m)$. The supervision of the student model is denoted as:
\begin{equation} \label{eq:Loss_LS}
    \setlength{\abovedisplayskip}{5pt} 
    \setlength{\belowdisplayskip}{5pt}
    \resizebox{0.56\hsize}{!}{$\begin{aligned}
        \mathcal{L}_{s} &= \lambda  \cdot\boldsymbol{s}(p,\mathcal{N}) \cdot  
        \mathcal{L}_{pxl}\left(\Theta^s\left(x_{D_2} \right),p \right) \\
        +& \lambda  \cdot \boldsymbol{s}(p,\mathcal{N}) \cdot 
	    \mathcal{L}_{pxl}\left(\Theta^s\left(x _{D_2}^{\prime}\right),{p }^{\prime}\right)
        \\
        +& (1-\lambda) \cdot \mathcal{L}_{cs}\left(\Theta^s\left(x_{D_2} \right)^{\prime},\Theta^s\left(x _{D_2}^{\prime}\right)\right)\\
        + & (1-\lambda)\cdot  \mathcal{L}_{pn}\left(\Theta^s\left(x_{D_2} \right),\mathcal{N}_ {m}\right) \\
        + & (1-\lambda) \cdot \mathcal{L}_{pn}\left(\Theta^s\left(x _{D_2}^{\prime}\right),\mathcal{N}_ {m}^{\prime}\right).
    \end{aligned}$}
\end{equation}
All the training samples,$x_{D_2}$, are preprocessed by PNMxp, PVRMxp in succession.

Algorithm~\ref{alg:A} summarizes the experimental procedure of PNL.

\begin{algorithm}[!htbp]
    \caption{\fontsize{8.5}{18}\selectfont The proposed Point-Neighborhood Learning method.}  
    \footnotesize 
    \label{alg:A}  
    \hspace*{0.02in}{\textbf{Input:}} Image $ x_{D_{1,2}} $; Ground truth of $D_1$; Point annotation of $D_{1,2}$; Radius $\mathcal{R} $.
    \begin{algorithmic}[1]
        \STATE // Generate the point-neighborhood bank $\mathcal{B_{\mathcal{N}}}$.
     
        \STATE // Train the teacher model $\Theta^t$ on $D_1$. $E_t$: epoch times.
    \FOR{ $e=1,2,...,E_t$}
        \FOR{ $x\in D_1 $}
        \STATE 1. $\widetilde{x}=$ PNMxp ($x$,$B_{\mathcal{PN}}$);
        \STATE 2. $\hat{y} = \Theta^t( \widetilde{x})$, $\hat{y}^{\prime} = \Theta^t( \widetilde{x}^{\prime})$;
        \STATE 3. Calculate the losse: $\mathcal{L}_t$;
        \STATE 4. Update the weight of $\Theta^t$;
        \ENDFOR
    \ENDFOR
    \STATE // Train the student $\Theta^s$ and update the teacher $ \Theta^t$. $E_s$: epoch times.
    \FOR{ $e=1,2,...,E_s$}
    \STATE Update the pseudo-labels on $D_2$ with $\Theta^t$, $p = \Theta^t\left(x_{D_2}\right) $.
        \FOR{ $m=1,2,...,M$}
            \FOR{ $x_{D_2}\in D_2 $}
            \STATE 1. Sample $x_{D_1}$ randomly from $D_1$
            \STATE 2. $\widetilde{x} $= PNMxp ($x$,$B_{\mathcal{N}}$);
            \STATE 2. $\widetilde{x}=$ PVRMxp ($\widetilde{x}_{D_1}$, $\widetilde{x}_{D_2}$);
            \STATE 3. $\hat{y} = \Theta^s( \widetilde{x})$, $\hat{y}^{\prime} = \Theta^s( \widetilde{x}^{\prime})$;
            \STATE 4. Calculate the losse: $\mathcal{L}_s$;
            \STATE 5. Update the weight of $\Theta^s$;
            \ENDFOR
            \STATE Update the weight of $\Theta^t$ with the EMA of $\Theta^s$;
        \ENDFOR
    \ENDFOR
    \end{algorithmic}  
    \hspace*{0.02in}{\textbf{Output:}} Updated weight of the teacher model $\Theta^t$.
\end{algorithm}

\section{Experiments and Results}
\label{sec_experiment} 
\subsection{Datasets}\label{subsec_dataset}
We evaluate our method using the \textit{NPC-LES} (2023) dataset, a private nasal endoscopic image dataset designed for nasopharyngeal carcinoma (NPC) lesion segmentation. The \textit{NPC-LES} dataset was collected from diagnostic practices at the First Affiliated Hospital of Sun Yat-sen University, with ethical approval and informed consent obtained from all participating patients. The dataset consists of 3,182 training samples, each annotated with both pixel-level and single-point annotations, as well as 453 test samples with pixel-level annotations only. To ensure subject independence between subsets $D_1$ and $D_2$, we partition the training set as follows: $D_1$ contains 151 images with both pixel-level and point annotations, accounting for 4.7\% of the training set, while $D_2$ consists of 3,031 images with only point annotations, accounting for 95.3\% of the training set.

The NPC lesion segmentation on nasal endoscopic images is a large object segmentation task. To verify both the generalization of our method and the validity of our point-neighborhood assumption, we additionally test our method on three segmentation tasks that involve relatively small objects: \textit{Kvasir-SEG} (2020)\cite{Kvasir_SEG}, \textit{CVC ClinicDB} (2015)\cite{CVC_ClinicDB_1, CVC_ClinicDB_2}, and \textit{ETIS Larib} (2014)\cite{ETIS_Larib}. These three datasets are all used for pixel-wise segmentation of colonic polyps. The \textit{Kvasir-SEG}\cite{Kvasir_SEG} dataset includes 1,000 endoscopic images with pixel-level annotations. The \textit{CVC ClinicDB}\cite{CVC_ClinicDB_1, CVC_ClinicDB_2} dataset contains 612 endoscopic images with corresponding pixel-level annotations. The \textit{ETIS Larib}\cite{ETIS_Larib} dataset comprises 196 endoscopic images with corresponding pixel-level masks. Following\cite{point_WangHong_MIA, sanderson2022fcn, nguyen2021ccbanet}, we split each dataset into a training set and a test set with an 80:20 ratio. We adopt the same strategy as\cite{point_WangHong_MIA} and\cite{point_OD_point_DETR} to generate single-point annotations for the training samples. Then, 10\% of the training samples are randomly selected as a pixel-level annotation subset, while the remaining 90\% are single-point annotation samples. The settings for the four datasets are shown in Table~\ref{Table:dataset}.

\begin{table}[ht]
    \centering
    \vspace{-0.4cm}
    \caption{Dataset configuration.}
    \belowrulesep=0pt
    \aboverulesep=0pt
    \begin{tabular}{c|c|ccc}
        \toprule
        \multirow{2}{*}{Dataset} & 
        \multirow{2}{*}{\makecell{Sample\\Number}} & 
        \multicolumn{2}{c}{Training Set} & 
        \multirow{2}{*}{Test Set}
        \\
        \cmidrule(lr){3-4} 
        & & \fontsize{7}{18}\selectfont pixel-level & \fontsize{7}{18}\selectfont point-level \\
        \midrule

        \makecell{\textit{NPC-LES} \fontsize{7}{18}\selectfont (Private)} & 3635 &  151 & 3031&453\\  
        \hline   
        \makecell{\textit{Kvasir-SEG} \fontsize{7}{18}\selectfont (Public)} & 1000 &  80 &720 &200\\ 
        \hline   
        \makecell{\textit{CVC ClinicDB} \fontsize{7}{18}\selectfont (Public)} & 612 &   49 &441 &122\\ 
        \hline   
        \makecell{\textit{ETIS Larib} \fontsize{7}{18}\selectfont (Public)} & 196 &   15 &142 &39\\ 
        \bottomrule
    \end{tabular}
    \label{Table:dataset}
    \vspace{-0.3cm}
\end{table}

\subsection{Implementation details}
In our comparison experiments with SOTA methods and ablation studies, the backbone network for PNL is set to DeepLabV3+\cite{backbone_DeepLabV3Plus}. The point-neighborhood hyperparameter, $\mathcal{R}$, is set to 20 pixels for all experiments across the four datasets. The EMA decaying factor, $\alpha$, is set to 0.995. The loss weight parameter, $\lambda$, is set to 0.8 when the teacher is pretrained and 0.5 during student model training. The number of epochs, $E_t$ and $E_s$, are set to 150 and 400, respectively.

\subsection{Evaluation Metrics} 

Five common segmentation metrics, Mean intersection over Union ($mIoU$), $Precision$, Pixel~Accuracy ($PA$), $Recall$ and $Dice$ as same as\cite{NPC_NiXiaoguang_2}, are used.
Intersection over Union (IoU) denotes the ratio between intersection and union of the segmentation prediction and the object.
Mean intersection over Union ($mIoU$) averages IoUs of foreground (lesion) and background (healthy tissue), which is defined as:
\begin{equation} \label{eq:mIoU}
    \resizebox{0.67\hsize}{!}{$\begin{aligned}
    mIoU = \frac{1}{k} \sum_{i=1}^{k} \frac{TP_i}{FN_i+FP_i+TP_i } \times 100\%,
    \end{aligned}$}
\end{equation}
where $ k $ is the total number of classes. $TP$, $FN$, $FP$ and $TN$ represent the true positive, false negative, false positive, and true negative predictions, respectively.
The $Precision$, $Recall$, $PA$ and $Dice$ are defined as:
\begin{equation} \label{eq:precision}
    \resizebox{0.48\hsize}{!}{$\begin{aligned}
    Precision = \frac{TP}{FP+TP } \times 100\%.
    \end{aligned}$}
\end{equation}
\begin{equation}\label{eq:recall}
    \resizebox{0.45\hsize}{!}{$\begin{aligned}
        Recall = \frac{TP}{TP + FN}  \times 100\%,
\end{aligned}$}
\end{equation}
\begin{equation} \label{eq:pa}
    \resizebox{0.55\hsize}{!}{$\begin{aligned}
    PA = \frac{TP + TN}{TP + TN + FP + FN} \times 100\%, 
\end{aligned}$}
\end{equation}
\begin{equation} \label{eq:dice}
    \resizebox{0.55\hsize}{!}{$\begin{aligned}
    Dice = \frac{2 \times TP}{2 \times TP + FP + FN}  \times 100\%.
\end{aligned}$}
\end{equation}

 
\subsection{Comparison with SOTA}
 
    \begin{table*}[ht]
        \vspace{-0.3cm}
        \belowrulesep=0pt
        \aboverulesep=0pt
        \caption{The quantitative validation results on \textit{NPC-LES} (Mean \text{\fontsize{6}{18}\selectfont{$\pm$Standard deviation}})
        }
        \label{Table:different_framework_NPC}
        \centering
        \setlength{\tabcolsep}{8pt}
        \begin{tabular*}{\hsize}{@{}@{\extracolsep{\fill}}c|cccccc}
        \toprule
        Datasets& 
        Methods& 
        $mIoU$ $ \left( \% \right) $& 
        $PA$ $ \left( \% \right) $ &  
        $Recall$ $ \left( \% \right) $ & 
        $Precision$ $ \left( \% \right) $ & 
        $Dice$$ \left( \% \right) $  \\
        \hline
        
        \multirow{7}{*}{\makecell{\textit{NPC-LES} \\(Private Dataset)}}
        &BCP\cite{semi_mixup_BCP} \textit{\fontsize{6}{18}\selectfont (CVPR 2023)}
            & 77.24\fontsize{6}{18}\selectfont$\pm$2.40 & 89.03\fontsize{6}{18}\selectfont$\pm$1.09 & 83.80\fontsize{6}{18}\selectfont$\pm$2.78 & 84.21\fontsize{6}{18}\selectfont$\pm$2.61 & 77.63\fontsize{6}{18}\selectfont$\pm$3.22\\
        &STT\cite{semi_double_tch_21} \textit{\fontsize{6}{18}\selectfont (NIPS 2024)}
            & 73.32\fontsize{6}{18}\selectfont$\pm$2.34 & 86.94\fontsize{6}{18}\selectfont$\pm$1.28 & 75.03\fontsize{6}{18}\selectfont$\pm$2.27 & 86.95\fontsize{6}{18}\selectfont$\pm$3.13 & 73.00\fontsize{6}{18}\selectfont$\pm$2.40\\
        &MCF\cite{semi_MCF} \textit{\fontsize{6}{18}\selectfont (CVPR 2023)}
            & 71.39\fontsize{6}{18}\selectfont$\pm$1.73 & 87.77\fontsize{6}{18}\selectfont$\pm$0.56 & 78.98\fontsize{6}{18}\selectfont$\pm$2.64 & 77.64\fontsize{6}{18}\selectfont$\pm$1.03 & 69.42\fontsize{6}{18}\selectfont$\pm$2.92\\
        &POL\cite{Yu_2022_CVPR} \textit{\fontsize{6}{18}\selectfont (CVPR 2022)}
            &66.57\fontsize{6}{18}\selectfont$\pm$0.62&85.66\fontsize{6}{18}\selectfont$\pm$0.53&83.13\fontsize{6}{18}\selectfont$\pm$0.94&66.42\fontsize{6}{18}\selectfont$\pm$1.00&62.90\fontsize{6}{18}\selectfont$\pm$1.05\\
        &Point-SEGTR\cite{point_WangHong_MIA} \textit{\fontsize{6}{18}\selectfont (MIA 2023)}
            & 73.64\fontsize{6}{18}\selectfont$\pm$1.56 & 82.08\fontsize{6}{18}\selectfont$\pm$1.26 & 79.54\fontsize{6}{18}\selectfont$\pm$1.91 & 82.00\fontsize{6}{18}\selectfont$\pm$2.00 & 73.80\fontsize{6}{18}\selectfont$\pm$2.57\\
        &SAM (Fine-tuning)\cite{point_SAM} \textit{\fontsize{6}{18}\selectfont (ICCV 2023)}
            & 69.67\fontsize{6}{18}\selectfont$\pm$0.98 & 82.71\fontsize{6}{18}\selectfont$\pm$1.43 & 72.21\fontsize{6}{18}\selectfont$\pm$0.80 & 80.41\fontsize{6}{18}\selectfont$\pm$2.35 & 62.88\fontsize{6}{18}\selectfont$\pm$1.11 \\
        &\textbf{Ours}
            & \textbf{82.84\fontsize{6}{18}\selectfont$\pm$0.44}& \textbf{92.28\fontsize{6}{18}\selectfont$\pm$0.36}& \textbf{85.68\fontsize{6}{18}\selectfont$\pm$0.62}& \textbf{90.84\fontsize{6}{18}\selectfont$\pm$1.07}& \textbf{84.80\fontsize{6}{18}\selectfont$\pm$1.37}\\
        \bottomrule
        \end{tabular*}
        \vspace{-0.3cm}

    \end{table*}

    \subsubsection{SOTA Methods}
    We compare our method with six SOTA methods. Point-SEGTR\cite{point_WangHong_MIA} is a recently proposed method for NPC lesion segmentation, which use the point coordinates as prompts to fuse image features. SAM\cite{point_SAM} is a popular prompt segmentation model and it can predict lesion semantics based on given point prompts. In the experiment, we fine-tune the SAM model on the $D_1$ subset and use the fine-tuned SAM to predict pseudo-labels for $D_2$ subset. The pseudo-labels are used to train another prompt-free SAM. Additionally, we choose three representative semi-supervised methods: BCP\cite{semi_mixup_BCP}, MCF\cite{semi_MCF}, STT\cite{semi_double_tch_21} and one weakly supervised method: POL\cite{Yu_2022_CVPR} for comparison.

    \subsubsection{Quantitative Analysis} 
    \begin{figure*}[ht]
        \centering
        \includegraphics[width=\textwidth]{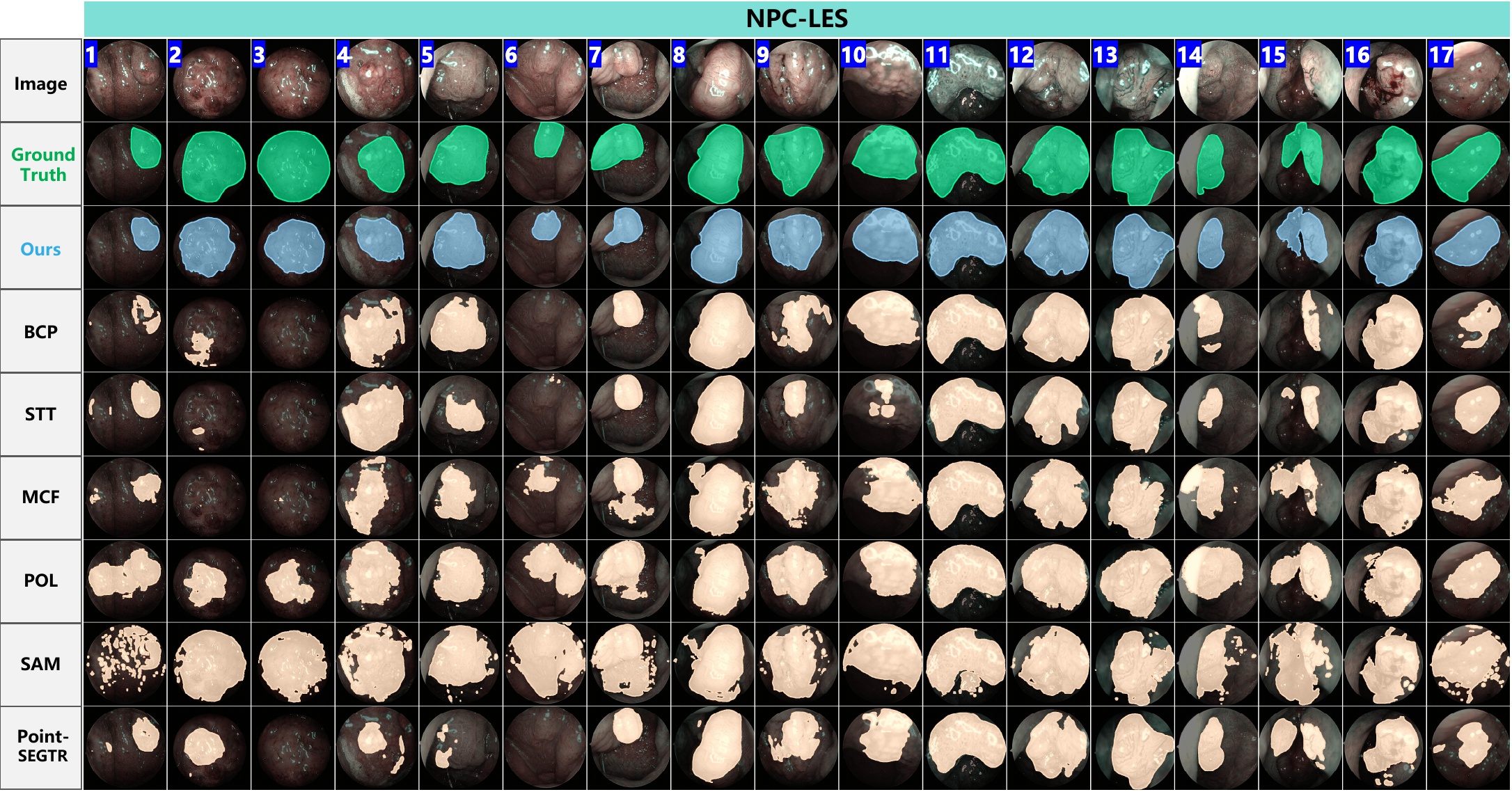}
        \caption{Visualization comparison of the predictions of different methods on the \textit{NPC-LES} dataset.}
        \label{fig:different_frameworks_NPC_LES}
        \vspace{-0.4cm}
    \end{figure*}

    Table~\ref{Table:different_framework_NPC} presents the results of various methods on the \textit{NPC-LES} dataset. It is clear that the proposed method outperforms all state-of-the-art (SOTA) approaches. BCP\cite{semi_mixup_BCP} has a natural advantage in large object segmentation, as its performance improves when the proportion of object areas is large. However, our method still significantly outperforms BCP, with improvements of 6.63\% in $Precision$ and 7.17\% in $Dice$. 
    STT\cite{semi_double_tch_21} improves performance by switching between different temporary teachers. Our method outperforms STT by 9.52\% in $mIoU$, 10.65\% in $Recall$, and 11.8\% in $Dice$. 
    MCF\cite{semi_MCF} relies on mutual correction between pixel-level labeled and pseudo-labeled images to improve the overall framework’s performance. However, if the object boundaries are blurred, the pseudo-labels may be biased, leading to incorrect corrections. This highlights the importance of high-quality pseudo-labels and underscores the value of our PSM approach.
    Due to the blurred boundaries, POL\cite{Yu_2022_CVPR} suffers in performance. Its $Recall$ metric (83.13\%) indicates that the lesions are predicted better, but its $mIoU$ and $Dice$ scores are lower (66.57\% and 62.90\%, respectively), suggesting that many false positives ($FP$) are predicted. 
    Point-SEGTR\cite{point_WangHong_MIA} achieves good segmentation compared to other semi-supervised and weakly supervised methods. However, our method significantly outperforms it across all metrics, with improvements of 9.2\% in $mIoU$, 10.2\% in $PA$, and 11.0\% in $Dice$. This difference can be attributed to the pseudo-labels used by Point-SEGTR\cite{point_WangHong_MIA} (see Section~\ref{subsubsec:pseudolabel}), where the point labels fail to impose strong constraints on the model during inference\cite{Zhou_2022_CVPR,10.1145/3560815}. Additionally, some foreground regions in the pseudo-labels do not contain the corresponding point labels (point annotations should appear inside the foreground). 
    In contrast, the proposed PNS and PSM explicitly enforce strong supervisory constraints between point annotations and predicted lesions, which greatly improves the quality of the pseudo-labels generated by our method. This, in turn, leads to significantly better final test performance.
    Among the methods, SAM\cite{point_SAM} shows a notable decline in performance, likely due to the increased difficulty of feature recognition for NPC lesions in the \textit{NPC-LES} dataset. Many lesion boundaries are ambiguous, and the distinction between lesions and non-lesions is relatively vague. This highlights SAM's poor ability to identify semantic boundaries with low edge strength\cite{SAM_not_perfect}.

    \subsubsection{Qualitative Analysis}

    Fig.~\ref{fig:different_frameworks_NPC_LES} shows the visualized test results of various methods on the \textit{NPC-LES} dataset. The lesions exhibit irregular shapes (e.g., samples 2, 9, 15), often have blurred boundaries (e.g., samples 1, 6, 9), and their internal features are similar to non-lesion areas (e.g., samples 1, 4, 6, 14, 17). Despite these challenges, our method successfully detects the lesions with complete regions and smooth boundaries, while other methods struggle with accurate segmentation.
    In more challenging cases (e.g., samples 2, 3, 6, 10), where the lesion contours are almost invisible and the internal features resemble non-lesion areas, both Point-SEGTR and our method detect the general lesion regions. However, our method is more precise in delineating the exact boundaries.
    These results validate the effectiveness of our proposed point-neighborhood learning strategy for learning from single-point weakly labeled data.

    \subsection{Experiments on public datasets with small objects}
    \begin{table*}[ht]
        \belowrulesep=0pt
        \aboverulesep=0pt
        \caption{The comparison validation on \textit{Kvasir-SEG}, \textit{CVC ClinicDB}, \textit{ETIS Larib} (Mean \text{\fontsize{6}{18}\selectfont$\pm$Standard deviation}).
        }
        \label{Table:different_framework_CVC_ETIS_Kvasir}
        \centering
        \setlength{\tabcolsep}{8pt}
        \begin{tabular*}{\hsize}{@{}@{\extracolsep{\fill}}c|cccccc}
        \toprule
        Datasets& 
        Methods& 
        $mIoU$ $ \left( \% \right) $& 
        $PA$ $ \left( \% \right) $ &  
        $Recall$ $ \left( \% \right) $ & 
        $Precision$ $ \left( \% \right) $ & 
        $Dice$$ \left( \% \right) $  \\
        \hline

        \multirow{7}{*}{\makecell{\textit{Kvasir-SEG} \\(Public Dataset)}}
        &BCP\cite{semi_mixup_BCP} \textit{\fontsize{6}{18}\selectfont (CVPR 2023)}
            &65.13\fontsize{6}{18}\selectfont$\pm$1.58&88.86\fontsize{6}{18}\selectfont$\pm$1.42&45.01\fontsize{6}{18}\selectfont$\pm$2.21&73.03\fontsize{6}{18}\selectfont$\pm$2.30&52.10\fontsize{6}{18}\selectfont$\pm$1.92 \\
        &STT\cite{semi_double_tch_21} \textit{\fontsize{6}{18}\selectfont (NIPS 2024)}
            &66.71\fontsize{6}{18}\selectfont$\pm$1.70&88.81\fontsize{6}{18}\selectfont$\pm$0.85&60.30\fontsize{6}{18}\selectfont$\pm$0.89&74.22\fontsize{6}{18}\selectfont$\pm$2.47&60.65\fontsize{6}{18}\selectfont$\pm$1.65\\
        &MCF\cite{semi_MCF} \textit{\fontsize{6}{18}\selectfont (CVPR 2023)}
            &66.75\fontsize{6}{18}\selectfont$\pm$1.25&88.04\fontsize{6}{18}\selectfont$\pm$1.16&65.53\fontsize{6}{18}\selectfont$\pm$1.36&68.02\fontsize{6}{18}\selectfont$\pm$1.36&60.24\fontsize{6}{18}\selectfont$\pm$1.55\\
        &POL\cite{Yu_2022_CVPR} \textit{\fontsize{6}{18}\selectfont (CVPR 2022)}
            &66.86\fontsize{6}{18}\selectfont$\pm$1.47&88.58\fontsize{6}{18}\selectfont$\pm$0.84&62.83\fontsize{6}{18}\selectfont$\pm$0.84&71.53\fontsize{6}{18}\selectfont$\pm$0.52&60.75\fontsize{6}{18}\selectfont$\pm$1.23\\
        &Point-SEGTR\cite{point_WangHong_MIA} \textit{\fontsize{6}{18}\selectfont (MIA 2023)}
            &78.89\fontsize{6}{18}\selectfont$\pm$0.54&92.35\fontsize{6}{18}\selectfont$\pm$0.35&83.14\fontsize{6}{18}\selectfont$\pm$0.87&77.94\fontsize{6}{18}\selectfont$\pm$1.10&77.48\fontsize{6}{18}\selectfont$\pm$1.13\\
        &SAM (Fine-tuning)\cite{point_SAM} \textit{\fontsize{6}{18}\selectfont (ICCV 2023)}
            &74.77\fontsize{6}{18}\selectfont$\pm$1.95&91.23\fontsize{6}{18}\selectfont$\pm$1.25&86.53\fontsize{6}{18}\selectfont$\pm$0.39&68.80\fontsize{6}{18}\selectfont$\pm$1.44&72.25\fontsize{6}{18}\selectfont$\pm$1.96\\
        &\textbf{Ours} &\textbf{83.19\fontsize{6}{18}\selectfont$\pm$1.01}&\textbf{94.38\fontsize{6}{18}\selectfont$\pm$0.43}&\textbf{80.00\fontsize{6}{18}\selectfont$\pm$1.44}&\textbf{91.32\fontsize{6}{18}\selectfont$\pm$0.42}&\textbf{81.76\fontsize{6}{18}\selectfont$\pm$0.77} \\
 
        \hline
        \multirow{7}{*}{\makecell{\textit{CVC ClinicDB} \\(Public Dataset)}}
        &BCP\cite{semi_mixup_BCP} \textit{\fontsize{6}{18}\selectfont (CVPR 2023)}
            &76.16\fontsize{6}{18}\selectfont$\pm$1.65 &95.72\fontsize{6}{18}\selectfont$\pm$1.16&60.33\fontsize{6}{18}\selectfont$\pm$2.11&92.51\fontsize{6}{18}\selectfont$\pm$2.61&66.65\fontsize{6}{18}\selectfont$\pm$1.31\\
        &STT\cite{semi_double_tch_21} \textit{\fontsize{6}{18}\selectfont (NIPS 2024)}
            &71.59\fontsize{6}{18}\selectfont$\pm$2.05&95.10\fontsize{6}{18}\selectfont$\pm$0.64&54.70\fontsize{6}{18}\selectfont$\pm$2.67&79.97\fontsize{6}{18}\selectfont$\pm$1.57&63.94\fontsize{6}{18}\selectfont$\pm$1.95\\
        &MCF\cite{semi_MCF} \textit{\fontsize{6}{18}\selectfont (CVPR 2023)}
            &75.20\fontsize{6}{18}\selectfont$\pm$1.59&95.49\fontsize{6}{18}\selectfont$\pm$1.18&59.69\fontsize{6}{18}\selectfont$\pm$2.05&88.66\fontsize{6}{18}\selectfont$\pm$1.76&66.35\fontsize{6}{18}\selectfont$\pm$1.40\\
        &POL\cite{Yu_2022_CVPR} \textit{\fontsize{6}{18}\selectfont (CVPR 2022)}
            &74.71\fontsize{6}{18}\selectfont$\pm$1.01&95.30\fontsize{6}{18}\selectfont$\pm$0.42&64.97\fontsize{6}{18}\selectfont$\pm$2.48&79.35\fontsize{6}{18}\selectfont$\pm$2.91&70.72\fontsize{6}{18}\selectfont$\pm$1.11\\
        &Point-SEGTR\cite{point_WangHong_MIA} \textit{\fontsize{6}{18}\selectfont (MIA 2023)}
            &79.29\fontsize{6}{18}\selectfont$\pm$0.39&95.97\fontsize{6}{18}\selectfont$\pm$0.81&69.30\fontsize{6}{18}\selectfont$\pm$0.77&83.62\fontsize{6}{18}\selectfont$\pm$1.21&75.98\fontsize{6}{18}\selectfont$\pm$1.04\\
        &SAM (Fine-tuning)\cite{point_SAM} \textit{\fontsize{6}{18}\selectfont (ICCV 2023)}
            &75.44\fontsize{6}{18}\selectfont$\pm$1.19&94.87\fontsize{6}{18}\selectfont$\pm$0.75&72.98\fontsize{6}{18}\selectfont$\pm$2.09&73.94\fontsize{6}{18}\selectfont$\pm$1.69&68.23\fontsize{6}{18}\selectfont$\pm$1.66\\
        &\textbf{Ours}      
            & \textbf{87.86\fontsize{6}{18}\selectfont$\pm$0.94}& \textbf{97.65\fontsize{6}{18}\selectfont$\pm$0.16}& \textbf{84.41\fontsize{6}{18}\selectfont$\pm$1.04}& \textbf{92.73\fontsize{6}{18}\selectfont$\pm$1.58}& \textbf{86.63\fontsize{6}{18}\selectfont$\pm$1.59}\\
        \hline

        \multirow{7}{*}{\makecell{\textit{ETIS Larib}\\(Public Dataset)}}
        &BCP\cite{semi_mixup_BCP} \textit{\fontsize{6}{18}\selectfont (CVPR 2023)}
            &51.71\fontsize{6}{18}\selectfont$\pm$1.29&96.22\fontsize{6}{18}\selectfont$\pm$1.68&~7.38\fontsize{6}{18}\selectfont$\pm$1.68&74.71\fontsize{6}{18}\selectfont$\pm$1.31&~8.34\fontsize{6}{18}\selectfont$\pm$1.35\\
        &STT\cite{semi_double_tch_21} \textit{\fontsize{6}{18}\selectfont (NIPS 2024)}
            &48.49\fontsize{6}{18}\selectfont$\pm$1.06&95.68\fontsize{6}{18}\selectfont$\pm$1.70&~1.32\fontsize{6}{18}\selectfont$\pm$1.32&42.50\fontsize{6}{18}\selectfont$\pm$2.26&9.689\fontsize{6}{18}\selectfont$\pm$2.36\\
        &MCF\cite{semi_MCF} \textit{\fontsize{6}{18}\selectfont (CVPR 2023)}
            &52.93\fontsize{6}{18}\selectfont$\pm$1.22&96.28\fontsize{6}{18}\selectfont$\pm$1.47&10.34\fontsize{6}{18}\selectfont$\pm$0.82&56.59\fontsize{6}{18}\selectfont$\pm$1.54&24.86\fontsize{6}{18}\selectfont$\pm$1.60\\
        &POL\cite{Yu_2022_CVPR} \textit{\fontsize{6}{18}\selectfont (CVPR 2022)}
            &50.20\fontsize{6}{18}\selectfont$\pm$1.14&88.93\fontsize{6}{18}\selectfont$\pm$1.59&40.89\fontsize{6}{18}\selectfont$\pm$2.32&31.23\fontsize{6}{18}\selectfont$\pm$3.00&37.38\fontsize{6}{18}\selectfont$\pm$2.91\\
        &Point-SEGTR\cite{point_WangHong_MIA} \textit{\fontsize{6}{18}\selectfont (MIA 2023)}
            &54.06\fontsize{6}{18}\selectfont$\pm$0.67&96.20\fontsize{6}{18}\selectfont$\pm$1.70&14.36\fontsize{6}{18}\selectfont$\pm$1.46&56.42\fontsize{6}{18}\selectfont$\pm$2.10&28.08\fontsize{6}{18}\selectfont$\pm$3.70\\
        &SAM (Fine-tuning)\cite{point_SAM} \textit{\fontsize{6}{18}\selectfont (ICCV 2023)}
            &48.11\fontsize{6}{18}\selectfont$\pm$0.43&95.90\fontsize{6}{18}\selectfont$\pm$0.96&0.355\fontsize{6}{18}\selectfont$\pm$0.24&37.49\fontsize{6}{18}\selectfont$\pm$0.80&50.90\fontsize{6}{18}\selectfont$\pm$0.13\\
        &\textbf{Ours}    &\textbf{68.15\fontsize{6}{18}\selectfont$\pm$1.26}&\textbf{97.13\fontsize{6}{18}\selectfont$\pm$0.67}&\textbf{54.46\fontsize{6}{18}\selectfont$\pm$1.00}&\textbf{66.61\fontsize{6}{18}\selectfont$\pm$0.94}&\textbf{60.68\fontsize{6}{18}\selectfont$\pm$1.06}\\ 
        \bottomrule
        \end{tabular*}
        \vspace{-0.4cm}

    \end{table*}

\begin{figure*}[!ht]
        \centering
        \vspace{-0.2cm}
        \includegraphics[width=\textwidth]{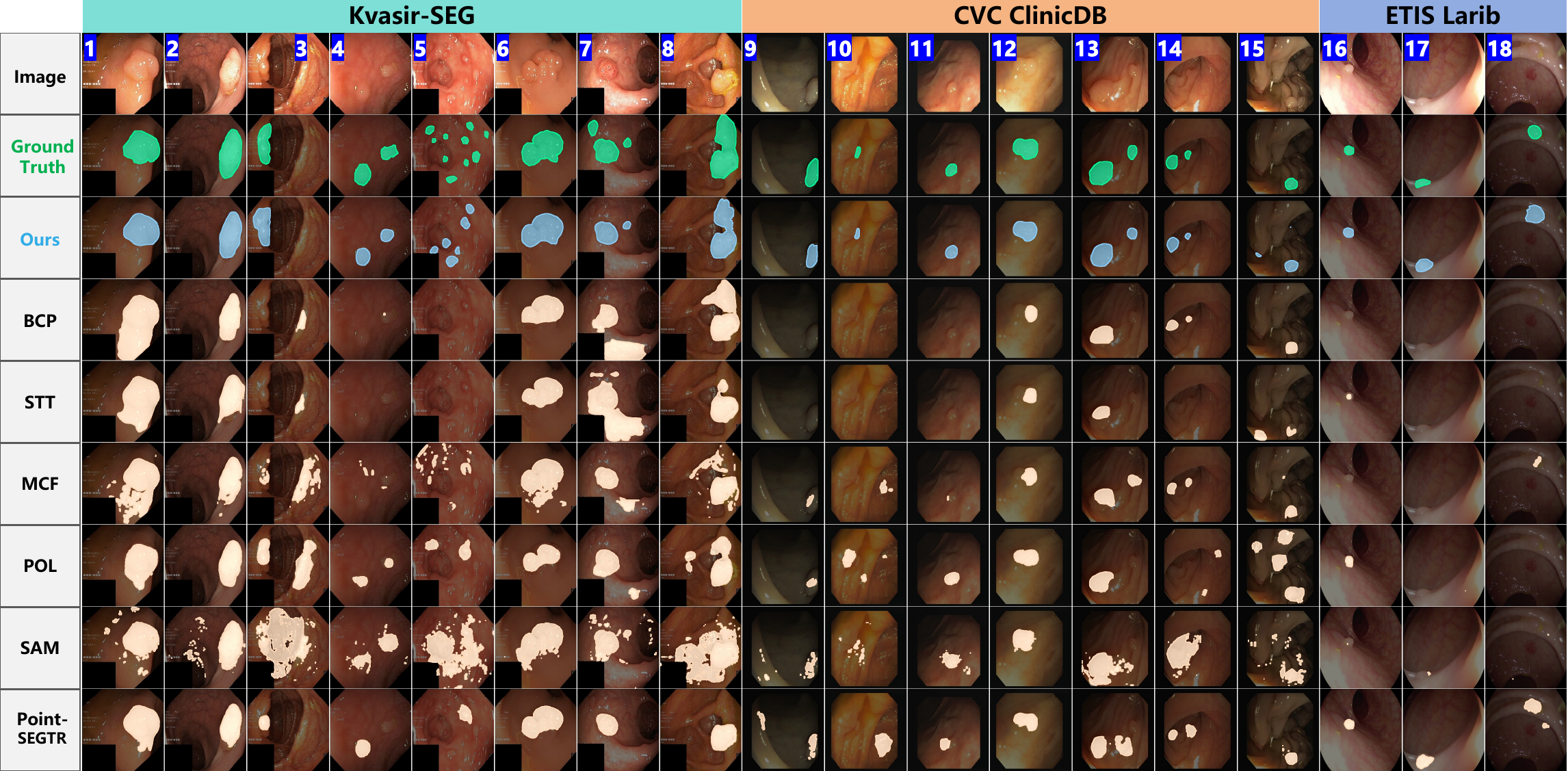}
        \caption{Visualization comparison of the predictions of different methods on the \textit{Kvasir-SEG}, \textit{CVC ClinicDB} and \textit{ETIS Larib} datasets.}
        \label{fig:different_frameworks_CVC_ETIS_Kvasir}
        \vspace{-0.4cm}
    \end{figure*}

    Table~\ref{Table:different_framework_CVC_ETIS_Kvasir} compares the performance of various methods on \textit{Kvasir-SEG}, \textit{CVC ClinicDB}, and \textit{ETIS Larib} datasets. Notably, our method demonstrates significant improvements in metrics like $Recall$ and $Precision$, which are crucial for small-object segmentation tasks. On the \textit{Kvasir-SEG} dataset, we achieve a 13.38\% increase in $Precision$, indicating our method's enhanced ability to reduce false positives ($FP$) and better discriminate small lesion areas. This is further evident on the \textit{CVC ClinicDB} and \textit{ETIS Larib} datasets, where the small size of the segmentation targets makes precision critical. Our approach excels in minimizing false negatives ($FN$), as reflected in the superior $Recall$ scores. These improvements directly correlate with our method's effectiveness in handling small objects, as it ensures accurate segmentation even in challenging cases. Fig.~\ref{fig:different_frameworks_CVC_ETIS_Kvasir} visually demonstrates these advancements in small-object scenarios (e.g., samples 9, 10, 11, 12, 13, 14, 16, 17 and 18).
    When the image contains multiple small lesions (e.g., samples 4, 5, 7, 13, and 14), our method effectively segments most of the polyps, while other methods struggle with recognizing multiple small objects. Additionally, our method excels in accurately delineating boundaries (e.g., samples 1 and 4).

\begin{table*}[h!]
    \belowrulesep=0pt
    \aboverulesep=0pt
    \centering
    \caption{Ablation Experiment on \textit{NPC-LES}, \textit{Kvasir-SEG}, \textit{CVC ClinicDB} and \textit{ETIS Larib} Datasets (Mean \text{\fontsize{6}{18}\selectfont$\pm$Standard deviation})}
    \label{Table:ablation_experiment}
    \begin{tabular*}{\hsize}{@{}@{\extracolsep{\fill}}c|cccc|ccccc}
    \toprule
    \multirow{2}{*}{Dataset} &
    \multicolumn{4}{c|}{Component} &
    \multirow{2}{*}{$mIoU$ $ \left( \% \right) $} &  
    \multirow{2}{*}{$PA$ $ \left( \% \right) $} &  
    \multirow{2}{*}{$Recall$ $ \left( \% \right) $} & 
    \multirow{2}{*}{$Precision$ $ \left( \% \right) $} & 
    \multirow{2}{*}{$Dice$ $ \left( \% \right) $}
    \\
     \cmidrule(lr){2-5}  
      & PNS & PSM & PNMxp & PVRMxp \\
    \hline
    \multirow{5}{*}{\makecell{\textit{NPC-LES} \\(Private Dataset)}}     
    & \usym{1F5F4}&\checkmark&\checkmark&  \checkmark  &75.13\fontsize{6}{18}\selectfont$\pm$1.31 &88.42\fontsize{6}{18}\selectfont$\pm$0.50 &84.66\fontsize{6}{18}\selectfont$\pm$0.95 &80.06\fontsize{6}{18}\selectfont$\pm$1.63 &74.49\fontsize{6}{18}\selectfont$\pm$1.96 \\ 
    & \checkmark &\usym{1F5F4}&  \checkmark&\checkmark  &77.09\fontsize{6}{18}\selectfont$\pm$1.76&89.33\fontsize{6}{18}\selectfont$\pm$1.02&76.96\fontsize{6}{18}\selectfont$\pm$1.98&84.94\fontsize{6}{18}\selectfont$\pm$1.71&77.65\fontsize{6}{18}\selectfont$\pm$1.76\\
    & \checkmark &\checkmark&  \usym{1F5F4}&  \checkmark  &76.33\fontsize{6}{18}\selectfont$\pm$1.59  &89.00\fontsize{6}{18}\selectfont$\pm$1.19&76.53\fontsize{6}{18}\selectfont$\pm$1.63&90.26\fontsize{6}{18}\selectfont$\pm$1.89&77.65\fontsize{6}{18}\selectfont$\pm$1.80\\
    & \checkmark &\checkmark&\checkmark&  \usym{1F5F4} &75.61\fontsize{6}{18}\selectfont$\pm$0.34&88.68\fontsize{6}{18}\selectfont$\pm$0.24&80.49\fontsize{6}{18}\selectfont$\pm$0.52&85.45\fontsize{6}{18}\selectfont$\pm$0.50&77.22\fontsize{6}{18}\selectfont$\pm$0.81\\
    & \checkmark &\checkmark&\checkmark&  \checkmark   & \textbf{82.84\fontsize{6}{18}\selectfont$\pm$0.44}& \textbf{92.28\fontsize{6}{18}\selectfont$\pm$0.36}& \textbf{85.68\fontsize{6}{18}\selectfont$\pm$0.62}& \textbf{90.84\fontsize{6}{18}\selectfont$\pm$1.07}& \textbf{84.80\fontsize{6}{18}\selectfont$\pm$1.37}\\
    \hline
    \multirow{5}{*}{\makecell{\textit{Kvasir-SEG} \\(Public Dataset)}} 
    & \usym{1F5F4}&\checkmark&\checkmark&  \checkmark  &74.87\fontsize{6}{18}\selectfont$\pm$1.29&91.42\fontsize{6}{18}\selectfont$\pm$1.69&65.66\fontsize{6}{18}\selectfont$\pm$1.36&90.09\fontsize{6}{18}\selectfont$\pm$0.59&69.95\fontsize{6}{18}\selectfont$\pm$0.91 \\ 
    & \checkmark &\usym{1F5F4}&  \checkmark&\checkmark &77.30\fontsize{6}{18}\selectfont$\pm$1.68&92.28\fontsize{6}{18}\selectfont$\pm$0.91&75.81\fontsize{6}{18}\selectfont$\pm$2.38&84.05\fontsize{6}{18}\selectfont$\pm$3.69&74.51\fontsize{6}{18}\selectfont$\pm$1.24 \\
    & \checkmark &\checkmark&  \usym{1F5F4}&  \checkmark  &78.42\fontsize{6}{18}\selectfont$\pm$0.90&92.15\fontsize{6}{18}\selectfont$\pm$0.78&78.21\fontsize{6}{18}\selectfont$\pm$1.01&74.56\fontsize{6}{18}\selectfont$\pm$1.24&72.60\fontsize{6}{18}\selectfont$\pm$0.66 \\
    & \checkmark &\checkmark&\checkmark&  \usym{1F5F4}  &69.70\fontsize{6}{18}\selectfont$\pm$2.08&88.73\fontsize{6}{18}\selectfont$\pm$2.12&68.22\fontsize{6}{18}\selectfont$\pm$3.52&75.35\fontsize{6}{18}\selectfont$\pm$4.74&65.79\fontsize{6}{18}\selectfont$\pm$0.65 \\
    & \checkmark &\checkmark&\checkmark&  \checkmark  &\textbf{83.19\fontsize{6}{18}\selectfont$\pm$1.01}&\textbf{94.38\fontsize{6}{18}\selectfont$\pm$0.43}&\textbf{80.00\fontsize{6}{18}\selectfont$\pm$1.44}&\textbf{91.32\fontsize{6}{18}\selectfont$\pm$0.42}&\textbf{81.76\fontsize{6}{18}\selectfont$\pm$0.77} \\
    \hline
    \multirow{5}{*}{\makecell{\textit{CVC ClinicDB} \\(Public Dataset)}} 
    & \usym{1F5F4}&\checkmark&\checkmark&  \checkmark  &81.49\fontsize{6}{18}\selectfont$\pm$1.04&96.68\fontsize{6}{18}\selectfont$\pm$0.82&68.53\fontsize{6}{18}\selectfont$\pm$1.71&95.14\fontsize{6}{18}\selectfont$\pm$1.89&77.79\fontsize{6}{18}\selectfont$\pm$1.00 \\ 
    & \checkmark &\usym{1F5F4}&  \checkmark&\checkmark  &82.77\fontsize{6}{18}\selectfont$\pm$1.18&96.52\fontsize{6}{18}\selectfont$\pm$0.09&72.44\fontsize{6}{18}\selectfont$\pm$1.85&95.54\fontsize{6}{18}\selectfont$\pm$0.75&78.71\fontsize{6}{18}\selectfont$\pm$1.28 \\
    & \checkmark &\checkmark&  \usym{1F5F4}&  \checkmark  &85.39\fontsize{6}{18}\selectfont$\pm$1.58&97.13\fontsize{6}{18}\selectfont$\pm$0.62&84.07\fontsize{6}{18}\selectfont$\pm$2.14&87.40\fontsize{6}{18}\selectfont$\pm$1.30&83.12\fontsize{6}{18}\selectfont$\pm$1.61 \\
    & \checkmark &\checkmark&\checkmark&  \usym{1F5F4}  &81.98\fontsize{6}{18}\selectfont$\pm$1.03&96.51\fontsize{6}{18}\selectfont$\pm$0.69&79.63\fontsize{6}{18}\selectfont$\pm$1.53&84.51\fontsize{6}{18}\selectfont$\pm$1.33&78.54\fontsize{6}{18}\selectfont$\pm$1.78 \\
    & \checkmark &\checkmark&\checkmark&  \checkmark  & \textbf{87.86\fontsize{6}{18}\selectfont$\pm$0.94}& \textbf{97.65\fontsize{6}{18}\selectfont$\pm$0.16}& \textbf{84.41\fontsize{6}{18}\selectfont$\pm$1.04}& \textbf{92.73\fontsize{6}{18}\selectfont$\pm$1.58}& \textbf{86.63\fontsize{6}{18}\selectfont$\pm$1.59}\\
    \hline
    \multirow{5}{*}{\makecell{\textit{ETIS Larib}\\(Public Dataset)}} 
    & \usym{1F5F4}&\checkmark&\checkmark&  \checkmark  &56.10\fontsize{6}{18}\selectfont$\pm$1.22&95.96\fontsize{6}{18}\selectfont$\pm$0.78&18.10\fontsize{6}{18}\selectfont$\pm$2.80&69.04\fontsize{6}{18}\selectfont$\pm$2.94&47.30\fontsize{6}{18}\selectfont$\pm$2.20 \\ 
    & \checkmark &\usym{1F5F4}&  \checkmark&\checkmark  &65.97\fontsize{6}{18}\selectfont$\pm$0.53&96.23\fontsize{6}{18}\selectfont$\pm$0.79&44.46\fontsize{6}{18}\selectfont$\pm$0.81&62.00\fontsize{6}{18}\selectfont$\pm$1.18&55.23\fontsize{6}{18}\selectfont$\pm$1.06 \\
    & \checkmark &\checkmark&  \usym{1F5F4}&  \checkmark  &54.68\fontsize{6}{18}\selectfont$\pm$1.56&83.44\fontsize{6}{18}\selectfont$\pm$1.60&37.03\fontsize{6}{18}\selectfont$\pm$0.77&38.11\fontsize{6}{18}\selectfont$\pm$1.90&43.45\fontsize{6}{18}\selectfont$\pm$2.36 \\
    & \checkmark &\checkmark&\checkmark&  \usym{1F5F4}  &54.31\fontsize{6}{18}\selectfont$\pm$0.81&86.34\fontsize{6}{18}\selectfont$\pm$1.92&39.14\fontsize{6}{18}\selectfont$\pm$1.24&27.50\fontsize{6}{18}\selectfont$\pm$3.54&38.13\fontsize{6}{18}\selectfont$\pm$2.85 \\
    & \checkmark &\checkmark&\checkmark&  \checkmark  &\textbf{68.15\fontsize{6}{18}\selectfont$\pm$1.26}&\textbf{97.13\fontsize{6}{18}\selectfont$\pm$0.67}&\textbf{54.46\fontsize{6}{18}\selectfont$\pm$1.00}&\textbf{66.61\fontsize{6}{18}\selectfont$\pm$0.94}&\textbf{60.68\fontsize{6}{18}\selectfont$\pm$1.06} \\
    \bottomrule
    \end{tabular*}
    \vspace{-0.2cm}
    \end{table*} 

\subsection{Ablation Study}
We conduct ablation studies to evaluate the contribution of each proposed component: Point-Neighborhood Supervision (PNS), Pseudo-Label Scoring Mechanism (PSM), Point-Neighborhood Mixup (PNMxp), and Particular-Value Randomly Mixup (PVRMxp). Single-component ablation experiments are performed across all four endoscopic image datasets, and the results are summarized in Table~\ref{Table:ablation_experiment}.

    \subsubsection{Effect of PNS}\label{subsubsec:point_neighborhood_supervision}

        To evaluate the effectiveness of the proposed PNS, we isolate it from the full framework ($\mathcal{FF}$), which includes PNS, PSM, PNMxp, and PVRMxp. PNS monitors the model’s ability to classify pixels within point-neighborhoods. When isolated, performance across the \textit{NPC-LES}, \textit{Kvasir-SEG}, \textit{CVC ClinicDB}, and \textit{ETIS Larib} datasets is significantly affected. Table~\ref{Table:ablation_experiment} shows the quantitative results from the ablation experiments, where on average, $mIoU$ decreases by 8.61\%, $Recall$ by 16.9\%, and $Dice$ by 11.09\%.
        Notably, $Precision$ is less impacted because it lacks the $FN$ term in its formula. The results suggest that the primary factor influencing performance is $FN$, and the decline in performance reflects a higher proportion of false negatives when PNS is removed. The role of PNS is to supervise pixels within the point-neighborhoods and prevent positive pixels from being misclassified as negative ($FN$).
        The $PA$ metric is less affected because its denominator includes all predicted terms ($TP$, $TN$, $FP$, $FN$), and a decrease in $FN$ causes corresponding increases in the other terms, resulting in minimal change. Although the change in $PA$ is small, it indicates an increase in the proportion of $TP$ and $TN$, showing that PNS improves the accuracy of the model’s predictions.

    \subsubsection{Effect of PSM}\label{subsubsec:pseudo_scoring}
    In the ablation experiments for PSM, we replaced $\boldsymbol{s}(p,\mathcal{N}_m)$ in Eq.~(\ref{eq:Loss_LS}) with a fixed score, 1.0. The results in Table~\ref{Table:ablation_experiment} show that isolating PSM significantly affects performance. On average, $mIoU$, $Recall$, and $Dice$ decrease by 4.73\%, 8.72\%, and 6.94\%, respectively, indicating that PSM is a crucial component of our method. Notably, $Recall$ and $Dice$ experience the largest decreases. This is mainly due to the fact that their numerators are solely composed of true positives ($TP$).
    The PSM is designed to assess the accuracy of pseudo-labels within the point-neighborhood. Positive samples within the point-neighborhood contribute to the $TP$ term. PSM suppresses pseudo-labels those fail to adequately recall the lesions in the point-neighborhoods, adaptively reducing the influence of pseudo-labels with low $Recall$ or $Dice$. The observed variations in the metrics align with the design of the PSM, further demonstrating that PSM helps the model more accurately segment lesion areas.


    \subsubsection{Effect of PNMxp}\label{subsubsec:abl_PNMxp}
    In the ablation experiments for PNMxp, we only performed PVRMxp before the original image entered the network model. As shown in Table~\ref{Table:ablation_experiment}, the average decrease of $mIoU$ is 6.81\%, $PA$ is 4.93\%, $Recall$ is 7.18\%, $Precision$ is 12.79\%, and $Dice$ is 9.26\%.
    The performance variation is obvious, which indicates that the model learns more reliable and richer features after PNMxp augmentation. 
    PNMxp significantly expands the sample space and improve the data diversity of samples. The experiments successfully verifies the effectiveness of PNMxp.

    \subsubsection{Effect of PVRMxp}\label{subsubsec:abl_PVRMxp}
    In the ablation experiments for PVRMxp, we only performed PNMxp to augment images. As shown in Table~\ref{Table:ablation_experiment}, the average decrease of $mIoU$ is 10.11\%, $PA$ is 5.30\%, $Recall$ is 9.27\%, $Precision$ is 17.17\%, and $Dice$ is 13.55\%. The advantage of PVRMxp is that each learning sample contains pixel-level annotation and pseudo-labels of the point-annotated sample. In the learning process, the model can compare the supervision of both of the two samples at all times, and then significantly improve the model's ability to find the reasonable boundary between the lesion and the non-lesion areas.

\section{Discussion}

\begin{table*}[ht] 
    \caption{
                The Generalization Experiment on Different Segmentation Models (Dataset: \textit{NPC-LES}) (Mean \text{\fontsize{6}{18}\selectfont$\pm$Standard deviation})
            }
    \label{Table:different_backbone}        
    \centering
    \setlength{\tabcolsep}{3pt}
    \begin{tabular}{p{70pt}<{\centering} p{60pt}<{\centering} p{65pt}<{\centering} p{65pt}<{\centering} p{65pt}<{\centering} p{65pt}<{\centering} p{65pt}<{\centering}}
    \toprule
    \multicolumn{2}{c}{Backbone} &
    $mIoU$ $ \left( \% \right) $& 
    $PA$ $ \left( \% \right) $ & 
    $Recall$ $ \left( \% \right) $ & 
    $Precision$ $ \left( \% \right) $ & 
    $Dice$ $ \left( \% \right) $ \\
    \hline
    \multirow{2}{*}{DeepLabV3+\cite{backbone_DeepLabV3Plus}} & without our PNL& 65.15\fontsize{6}{18}\selectfont$\pm$1.35   & 85.12\fontsize{6}{18}\selectfont$\pm$1.29   & 77.12\fontsize{6}{18}\selectfont$\pm$2.50   & 68.88\fontsize{6}{18}\selectfont$\pm$2.58   & 60.75\fontsize{6}{18}\selectfont$\pm$2.27     \\
    & \textbf{with our PNL} & \textbf{82.84\fontsize{6}{18}\selectfont$\pm$0.44}   & \textbf{92.28\fontsize{6}{18}\selectfont$\pm$0.36}   & \textbf{85.68\fontsize{6}{18}\selectfont$\pm$0.62}   & \textbf{90.84\fontsize{6}{18}\selectfont$\pm$1.07}   & \textbf{84.80\fontsize{6}{18}\selectfont$\pm$1.37}    \\
    \hline
    \multirow{2}{*}{PSPNet\cite{backbone_PSPNet}}    &    without our PNL      & 67.31\fontsize{6}{18}\selectfont$\pm$1.98   & 85.09\fontsize{6}{18}\selectfont$\pm$1.06   & 76.61\fontsize{6}{18}\selectfont$\pm$2.58   & 74.57\fontsize{6}{18}\selectfont$\pm$3.61   & 65.53\fontsize{6}{18}\selectfont$\pm$3.06    \\
    &\textbf{with our PNL} & \textbf{75.99\fontsize{6}{18}\selectfont$\pm$1.52}   & \textbf{88.66\fontsize{6}{18}\selectfont$\pm$0.94}   & \textbf{76.13\fontsize{6}{18}\selectfont$\pm$1.75}   & \textbf{90.11\fontsize{6}{18}\selectfont$\pm$2.59}   & \textbf{77.88\fontsize{6}{18}\selectfont$\pm$2.09}    \\
    \hline
    \multirow{2}{*}{SegNet\cite{backbone_Segnet}}       &    without our PNL   & 69.75\fontsize{6}{18}\selectfont$\pm$2.46   & 86.72\fontsize{6}{18}\selectfont$\pm$1.24   & 79.61\fontsize{6}{18}\selectfont$\pm$4.29   & 75.45\fontsize{6}{18}\selectfont$\pm$3.55   & 67.08\fontsize{6}{18}\selectfont$\pm$3.35     \\
    &\textbf{with our PNL}  & \textbf{79.77\fontsize{6}{18}\selectfont$\pm$1.53}   & \textbf{90.76\fontsize{6}{18}\selectfont$\pm$0.74}   & \textbf{84.38\fontsize{6}{18}\selectfont$\pm$1.67}   & \textbf{87.38\fontsize{6}{18}\selectfont$\pm$0.95}   & \textbf{81.45\fontsize{6}{18}\selectfont$\pm$1.76}    \\
    \hline
    \multirow{2}{*}{SegFormer [\enquote{B2}]\cite{backbone_segformer}}       &    without our PNL   
    &72.44\fontsize{6}{18}\selectfont$\pm$1.57&88.99\fontsize{6}{18}\selectfont$\pm$0.79&86.91\fontsize{6}{18}\selectfont$\pm$1.37&71.96\fontsize{6}{18}\selectfont$\pm$1.24&68.28\fontsize{6}{18}\selectfont$\pm$0.57     \\
    &\textbf{with our PNL}  
    &\textbf{82.57\fontsize{6}{18}\selectfont$\pm$0.86}&\textbf{92.55\fontsize{6}{18}\selectfont$\pm$0.39}&\textbf{88.18\fontsize{6}{18}\selectfont$\pm$0.46}&\textbf{87.76\fontsize{6}{18}\selectfont$\pm$1.00}&\textbf{83.49\fontsize{6}{18}\selectfont$\pm$0.84}    \\
    \bottomrule
    \end{tabular}
    \vspace{-0.3cm}
\end{table*} 

In this section, we also validate the our method's generalization ability across different segmentation models, compare our method with fully supervised approaches, and assess the improvement of pseudo-labels. Notably, we conduct robustness experiments on point-neighborhood transformations, exploring the impact of the hyperparameter $\mathcal{R}$ on performance. The following are the recordings and analyses of these experiments.

\begin{figure}[ht]
    \centering
    \includegraphics[width=\columnwidth]{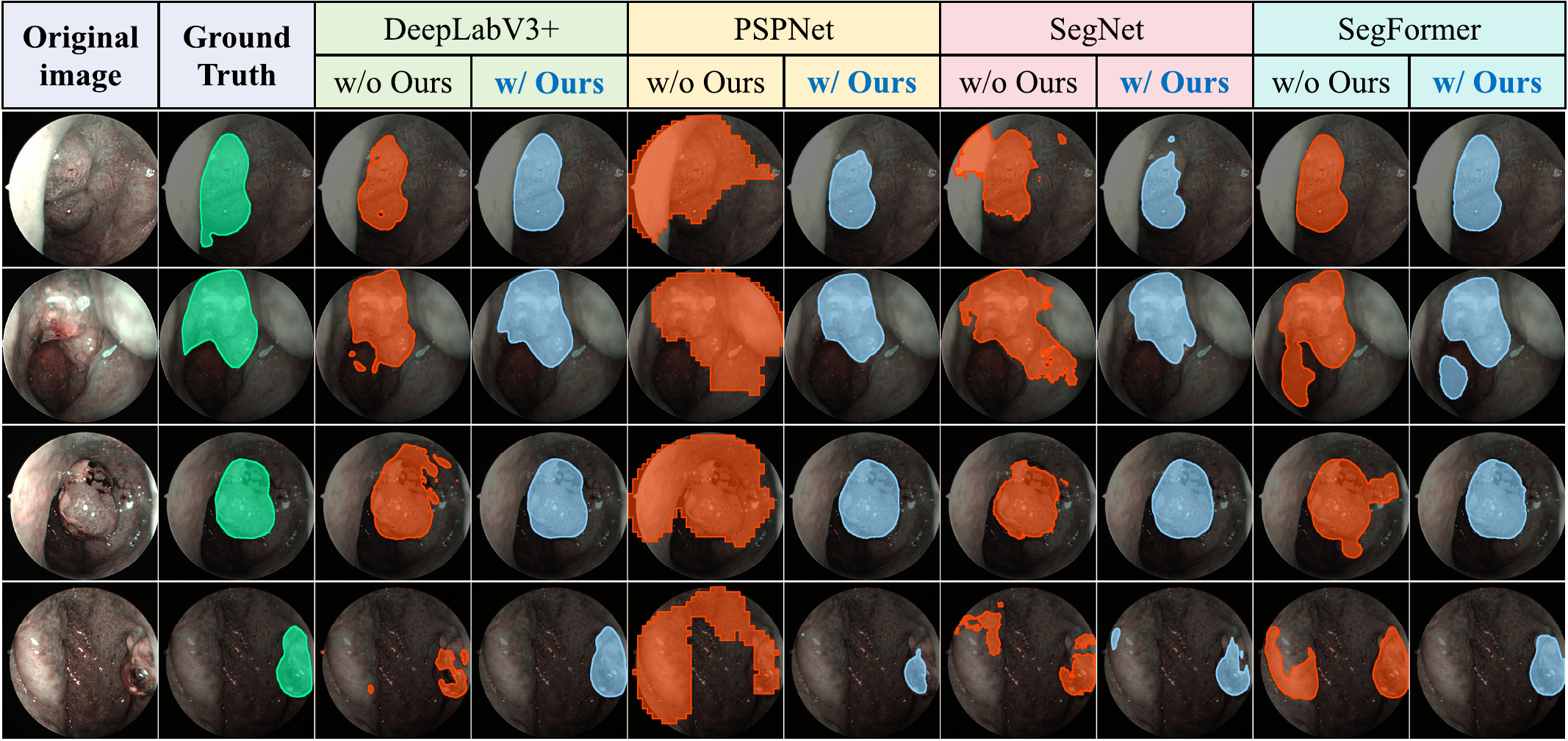}
    \caption{The visualization of improvements of our method on different segmentation networks (DeepLabV3+, PSPNet, SegNet and SegFormer [\enquote{B2}]).}
    \label{fig:different_backbone}
    \vspace{-0.3cm}
\end{figure}
\subsection{Generalization Ability of PNL}\label{subsubsec:different_backbone}
We validate our PNL on various segmentation networks: DeepLabV3+\cite{backbone_DeepLabV3Plus}, PSPNet\cite{backbone_PSPNet}, SegNet\cite{backbone_Segnet}, and SegFormer\cite{backbone_segformer} using the \textit{NPC-LES} dataset. The results, summarized in Table~\ref{Table:different_backbone} and visualized in Fig.~\ref{fig:different_backbone}, demonstrate that our method significantly improves generalization across all networks.
For DeepLabV3+, PNL enhances $mIoU$ by 17.69\%, $Precision$ by 21.96\%, and $Dice$ by 24.05\%. Similar improvements are observed across other models: PSPNet sees increases of 15.54\% in $Precision$ and 12.35\% in $Dice$; SegNet improves by 10.02\% in $mIoU$, 11.93\% in $Precision$, and 14.37\% in $Dice$; SegFormer benefits from 10.13\% in $mIoU$, 15.8\% in $Precision$, and 15.21\% in $Dice$. These results confirm the robustness and generalization ability of our method across different segmentation models.

\subsection{comparison with Fully-Supervised Learning (FSL) Method}

We conduct experiments to evaluate our method’s performance relative to fully-supervised learning (FSL) on four datasets, where all training samples have corresponding pixel-level annotations. As shown in Table~\ref{Table:Ideal_exp}, our method achieves results that are nearly on par with FSL, with only a small performance gap remaining in most cases. For example, on \textit{NPC-LES} and \textit{Kvasir-SEG}, the differences in metrics such as $mIoU$, $PA$, and $Precision$ are minimal, demonstrating that our method is effective even with limited supervision.
However, on the \textit{CVC ClinicDB} and \textit{ETIS Larib} datasets, the gap between our method and FSL is more pronounced. This is primarily due to the non-independent nature of their training and test sets. Specifically, in these datasets, there is a significant overlap between the training and test samples, which leads to data leakage and gives the FSL models an unfair advantage during testing. This overlap occurs because a large number of pixel-level annotated samples from the training set may already contain information that directly relates to the test set, thereby reducing the challenge of the segmentation task for the FSL models. In contrast, our method does not benefit from this leakage, resulting in a slightly larger gap in performance on these datasets.
Despite this, our approach still outperforms previous state-of-the-art methods, demonstrating its robustness and effectiveness in handling segmentation tasks even in the presence of challenging data conditions.

\begin{table}[!h]
    \centering
    \vspace{-0.2cm}
    \caption{comparison with the fully-supervised learning (FSL) method.}
    \setlength{\tabcolsep}{2pt}
    \begin{tabular}{ccccccc}
        \toprule
        Dataset&Mode   & $mIoU$           & $PA$   & $Recall$        &      $Precision$ &      $Dice$   \\
        \hline   
        \multirow{2}{*}{\makecell{\textit{NPC} \\\textit{-LES}}}
        &Ours       & 82.84\fontsize{6}{18}\selectfont$\pm$0.44   &  92.28\fontsize{6}{18}\selectfont$\pm$0.36  & 85.68\fontsize{6}{18}\selectfont$\pm$0.62  &      90.84\fontsize{6}{18}\selectfont$\pm$1.07&84.80\fontsize{6}{18}\selectfont$\pm$1.37\\
        &FSL      & 84.35\fontsize{6}{18}\selectfont$\pm$0.40   &  93.38\fontsize{6}{18}\selectfont$\pm$0.93  & 87.83\fontsize{6}{18}\selectfont$\pm$1.78  &      90.91\fontsize{6}{18}\selectfont$\pm$2.02&86.01\fontsize{6}{18}\selectfont$\pm$0.84\\
        \hline   
        \multirow{2}{*}{\makecell{\textit{Kvasir} \\\textit{-SEG}}}
        &Ours       & 83.19\fontsize{6}{18}\selectfont$\pm$1.01& 94.38\fontsize{6}{18}\selectfont$\pm$0.43& 80.00\fontsize{6}{18}\selectfont$\pm$1.44& 91.32\fontsize{6}{18}\selectfont$\pm$0.42& 81.76\fontsize{6}{18}\selectfont$\pm$0.77\\
        &FSL      &84.50\fontsize{6}{18}\selectfont$\pm$1.84&94.95\fontsize{6}{18}\selectfont$\pm$0.51&83.16\fontsize{6}{18}\selectfont$\pm$1.40&90.32\fontsize{6}{18}\selectfont$\pm$1.72&82.92\fontsize{6}{18}\selectfont$\pm$1.13\\
        \hline   
        \multirow{2}{*}{\makecell{\textit{CVC} \\\textit{ClinicDB}}}
        &Ours       & 87.86\fontsize{6}{18}\selectfont$\pm$0.94& 97.65\fontsize{6}{18}\selectfont$\pm$0.16& 84.41\fontsize{6}{18}\selectfont$\pm$1.04& 92.73\fontsize{6}{18}\selectfont$\pm$1.58& 86.63\fontsize{6}{18}\selectfont$\pm$1.59\\
        &FSL      &92.87\fontsize{6}{18}\selectfont$\pm$1.44&98.96\fontsize{6}{18}\selectfont$\pm$0.08&91.32\fontsize{6}{18}\selectfont$\pm$2.29&94.13\fontsize{6}{18}\selectfont$\pm$2.00&92.87\fontsize{6}{18}\selectfont$\pm$1.51\\
        \hline   
        \multirow{2}{*}{\makecell{\textit{ETIS} \\\textit{Larib}}}
        &Ours       &68.15\fontsize{6}{18}\selectfont$\pm$1.26&97.13\fontsize{6}{18}\selectfont$\pm$0.67&54.46\fontsize{6}{18}\selectfont$\pm$1.00&66.61\fontsize{6}{18}\selectfont$\pm$0.94&60.68\fontsize{6}{18}\selectfont$\pm$1.06\\
        &FSL      &77.85\fontsize{6}{18}\selectfont$\pm$1.54&98.70\fontsize{6}{18}\selectfont$\pm$0.43&60.70\fontsize{6}{18}\selectfont$\pm$0.89&85.34\fontsize{6}{18}\selectfont$\pm$1.12&71.60\fontsize{6}{18}\selectfont$\pm$0.97\\
        \bottomrule
    \end{tabular}
    \label{Table:Ideal_exp}
    \vspace{-0.4cm}
\end{table}

\subsection{Analysis on the Pseudo-labels}\label{subsubsec:pseudolabel}
\begin{figure}[h] 
    \centering
    \includegraphics[width=\columnwidth]{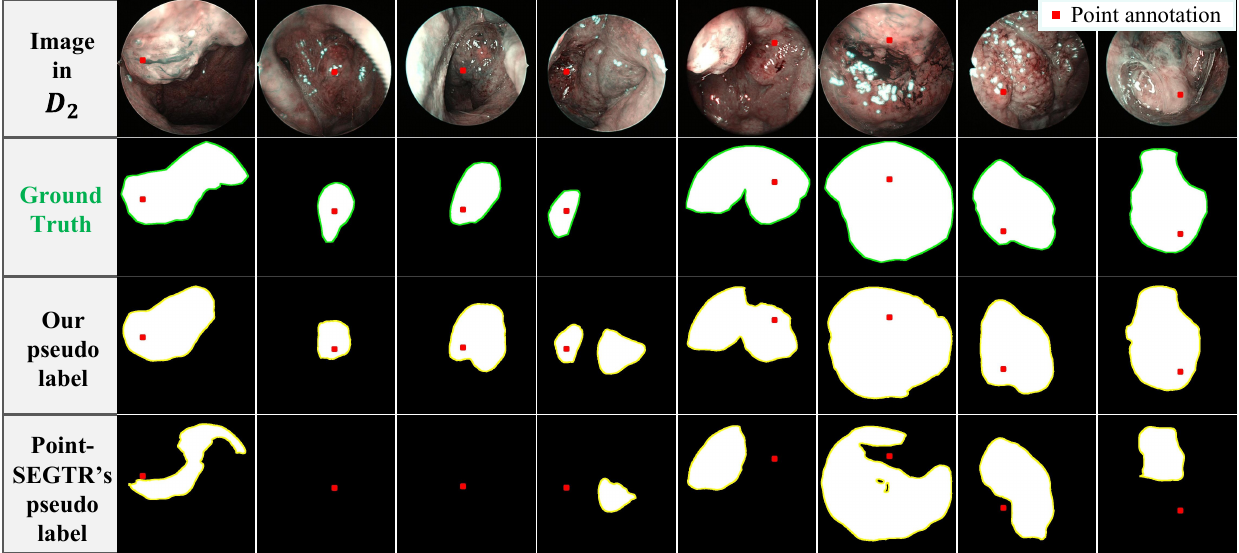}
    \caption{The visualization of the comparison of the pseudo-labels produced by Point-SEGTR\cite{point_WangHong_MIA} and our method.}
    \label{fig:point_value}
    \vspace{-0.1cm}
\end{figure}

The quality of pseudo-labels directly reflects the effectiveness of our method in learning from weakly semi-annotated data. As shown in Fig.~\ref{fig:point_value}, our pseudo-labels are more complete and consistent with the ground truth. Statistically, 99.83\% of the foreground areas in our pseudo-labels contain point annotations, compared to only 68.2\% in those generated by Point-SEGTR\cite{point_WangHong_MIA}. Furthermore, the $mIoU$ of our pseudo-labels reaches 81.41\%, while Point-SEGTR's $mIoU$ is only 72.49\%. These results demonstrate that our point-neighborhood strategies significantly improve the quality of pseudo-labels and enhance model performance.

\begin{figure}[h] 
    \centering 
    \includegraphics[height=3.5cm]{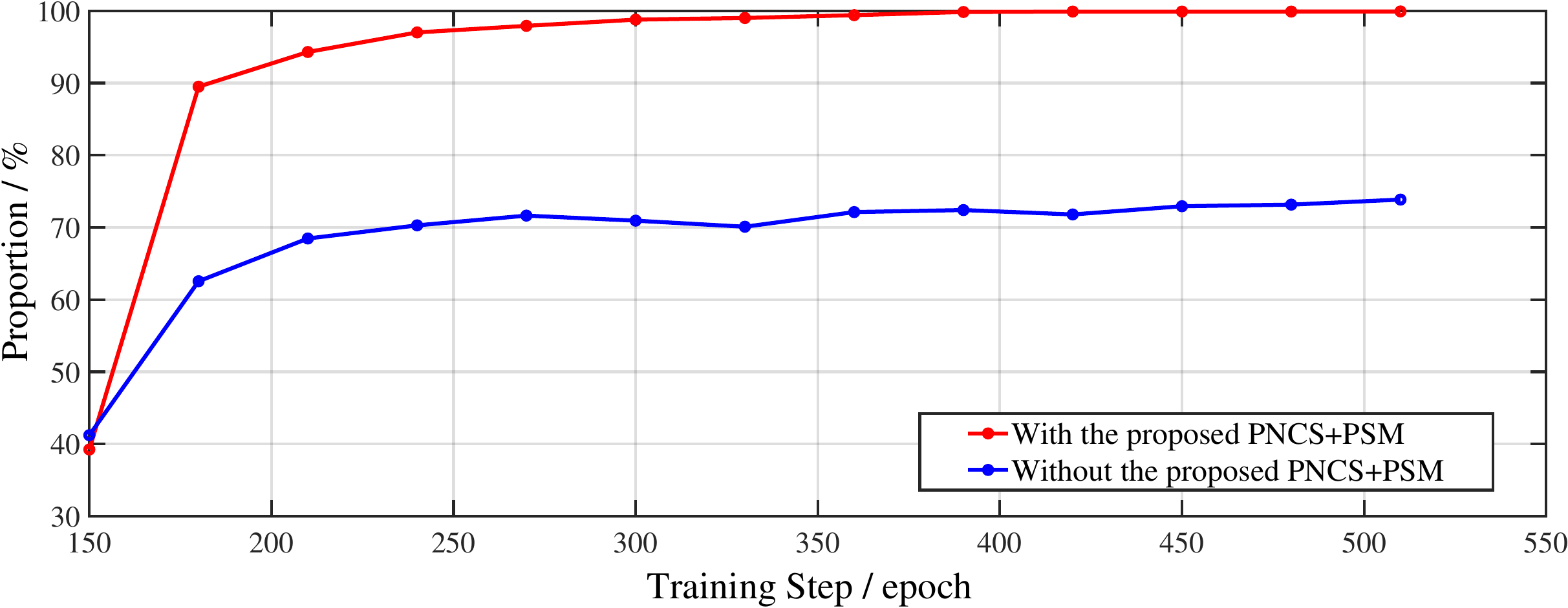}
    \caption{The proportion of pseudo-labels those contain the point annotations during training.}
    \label{fig:pseudo_label_point}
    \vspace{-0.3cm}
\end{figure}

We conducted additional experiments to assess the impact of PNS and PSM on the quality of pseudo-labels. Two sets of experiments were performed: the first (red line in Fig.~\ref{fig:pseudo_label_point}) uses our complete method ($\mathcal{FF}$), and the second (blue line in Fig.~\ref{fig:pseudo_label_point}) ablates PNS and PSM. We recorded the proportion of samples with annotated points inside the pseudo-labels. 
The results show that with PNS and PSM, the model quickly identifies point-annotated regions as lesions and converges faster. This demonstrates that our method effectively leverages the value of point annotations.


 
\subsection{Effect of hyperparameter $\mathcal{R}$ on segmentation performance}
\begin{figure}[h] 
    \centering
    \includegraphics[height=3.5cm]{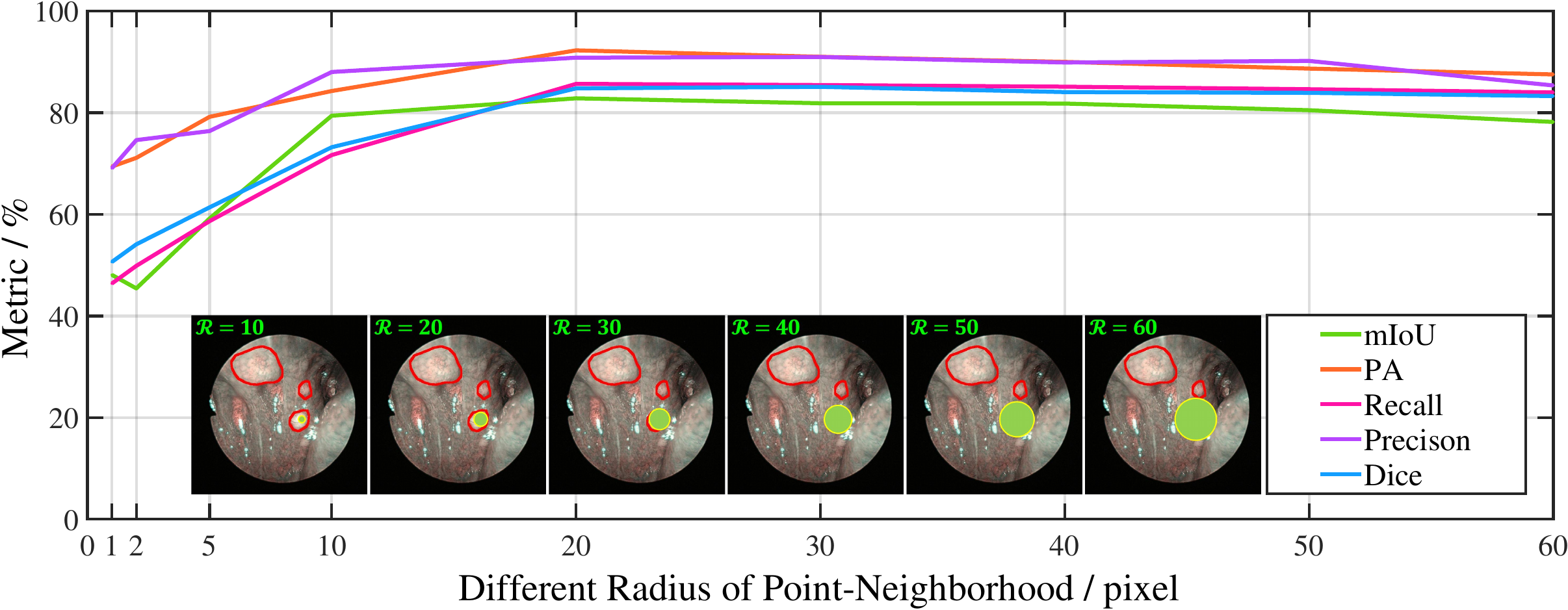}
    \caption{Curves of the segmentation metrics under different radius $\mathcal{R}$ of $\mathcal{N}$.}
    \label{fig:different_R}
    \vspace{-0.6cm}
\end{figure}

In previous experiments, the point-neighborhood hyperparameter was set to $\mathcal{R}=20$. In the NPC lesion segmentation task, point-neighborhoods are typically contained within the lesion. However, if $\mathcal{R}$ is too large and includes many non-lesion pixels, segmentation performance may degrade. To assess this, we conducted a robustness experiment on the NPC-LES dataset samples with a resolution of 512×512 pixels to test different values of $\mathcal{R} \in [1,60]$. As $\mathcal{R}$ increases, point-neighborhoods are increasingly contaminated with non-lesion pixels.
The performance of PNL for different $\mathcal{R}$ values is shown in Fig.~\ref{fig:different_R}. The results indicate that as point-neighborhoods expand beyond the lesion area, the model remains robust. However, when the neighborhoods shrink too small, performance is significantly impaired. This is because smaller neighborhoods capture fewer relevant lesion pixels, which limits the model's ability to effectively learn and segment the lesion areas.

\section{Conclusion}
\label{sec_Conclusion}
This paper presents an explicit yet efficient weakly semi-supervised segmentation method, Point-Neighborhood Learning (PNL). Our method leverages the label properties and reliable neighboring pixel information of point annotations, transforming them into point-neighborhood annotations. Key components of our method include PNS for supervision, PSM for suppressing low-quality pseudo-labels, and PNMxp and PVRMxp for data augmentation. Comprehensive experiments demonstrate that our method significantly outperforms SOTA methods.
Our future work will focus on exploring methods for few-shot weakly semi-supervised lesion segmentation tasks.

\bibliographystyle{IEEEtran}
\bibliography{ref}

\begin{thebibliography}{10}
\providecommand{\url}[1]{#1}
\csname url@samestyle\endcsname
\providecommand{\newblock}{\relax}
\providecommand{\bibinfo}[2]{#2}
\providecommand{\BIBentrySTDinterwordspacing}{\spaceskip=0pt\relax}
\providecommand{\BIBentryALTinterwordstretchfactor}{4}
\providecommand{\BIBentryALTinterwordspacing}{\spaceskip=\fontdimen2\font plus
\BIBentryALTinterwordstretchfactor\fontdimen3\font minus
  \fontdimen4\font\relax}
\providecommand{\BIBforeignlanguage}[2]{{%
\expandafter\ifx\csname l@#1\endcsname\relax
\typeout{** WARNING: IEEEtran.bst: No hyphenation pattern has been}%
\typeout{** loaded for the language `#1'. Using the pattern for}%
\typeout{** the default language instead.}%
\else
\language=\csname l@#1\endcsname
\fi
#2}}
\providecommand{\BIBdecl}{\relax}
\BIBdecl

\bibitem{NPC_review_TransBionologyReview}
Z.~Wang, M.~Fang, J.~Zhang, L.~Tang, L.~Zhong, H.~Li, R.~Cao, X.~Zhao, S.~Liu,
  R.~Zhang, X.~Xie, H.~Mai, S.~Qiu, J.~Tian, and D.~Dong, ``Radiomics and deep
  learning in nasopharyngeal carcinoma: A review,'' \emph{IEEE Reviews in
  Biomedical Engineering}, vol.~17, pp. 118--135, 2024.

\bibitem{NPC_MOHAMMED_4}
M.~A. Mohammed, M.~K. Abd~Ghani, N.~Arunkumar, R.~I. Hamed, S.~A. Mostafa,
  M.~K. Abdullah, and M.~Burhanuddin, ``Decision support system for
  nasopharyngeal carcinoma discrimination from endoscopic images using
  artificial neural network,'' \emph{The Journal of Supercomputing}, vol.~76,
  pp. 1086--1104, 2020.

\bibitem{NPC_MOHAMMED_5}
M.~K. Abd~Ghani, M.~A. Mohammed, N.~Arunkumar, S.~A. Mostafa, D.~A. Ibrahim,
  M.~K. Abdullah, M.~M. Jaber, E.~Abdulhay, G.~Ramirez-Gonzalez, and
  M.~Burhanuddin, ``Decision-level fusion scheme for nasopharyngeal carcinoma
  identification using machine learning techniques,'' \emph{Neural Computing
  and Applications}, vol.~32, pp. 625--638, 2020.

\bibitem{NPC_NiXiaoguang_1}
J.~Xu, J.~Wang, X.~Bian, J.-Q. Zhu, C.-W. Tie, X.~Liu, Z.~Zhou, X.-G. Ni, and
  D.~Qian, ``Deep learning for nasopharyngeal carcinoma identification using
  both white light and narrow-band imaging endoscopy,'' \emph{The
  Laryngoscope}, vol. 132, no.~5, pp. 999--1007, 2022.

\bibitem{NPC_NiXiaoguang_2}
S.-X. Wang, Y.~Li, J.-Q. Zhu, M.-L. Wang, W.~Zhang, C.-W. Tie, G.-Q. Wang, and
  X.-G. Ni, ``The detection of nasopharyngeal carcinomas using a neural network
  based on nasopharyngoscopic images,'' \emph{The Laryngoscope}, vol. 134,
  no.~1, pp. 127--135, 2024.

\bibitem{NPC_Li2018}
C.~Li, B.~Jing, L.~Ke, B.~Li, W.~Xia, C.~He, C.~Qian, C.~Zhao, H.~Mai, M.~Chen,
  K.~Cao, H.~Mo, L.~Guo, Q.~Chen, L.~Tang, W.~Qiu, Y.~Yu, H.~Liang, X.~Huang,
  G.~Liu, W.~Li, L.~Wang, R.~Sun, X.~Zou, S.~Guo, P.~Huang, D.~Luo, F.~Qiu,
  Y.~Wu, Y.~Hua, K.~Liu, S.~Lv, J.~Miao, Y.~Xiang, Y.~Sun, X.~Guo, and X.~Lv,
  ``Development and validation of an endoscopic images-based deep learning
  model for detection with nasopharyngeal malignancies,'' \emph{Cancer
  Communications}, vol.~38, no.~1, p.~59, 2018.

\bibitem{ganeshan2024enhancing}
V.~Ganeshan, J.~Bidwell, D.~Gyawali, T.~S. Nguyen, J.~Morse, M.~P. Smith, B.~M.
  Barton, and E.~D. McCoul, ``Enhancing nasal endoscopy: Classification,
  detection, and segmentation of anatomic landmarks using a convolutional
  neural network,'' \emph{International Forum of Allergy \& Rhinology},
  vol.~14, no.~9, pp. 1521--1524, 2024.

\bibitem{Xu2024}
Y.~Xu, J.~Wang, C.~Li, Y.~Su, H.~Peng, L.~Guo, S.~Lin, J.~Li, and D.~Wu,
  ``Advancing precise diagnosis of nasopharyngeal carcinoma through
  endoscopy-based radiomics analysis,'' \emph{iScience}, vol.~27, no.~9, Sep
  2024.

\bibitem{weak_CAM}
B.~Zhou, A.~Khosla, A.~Lapedriza, A.~Oliva, and A.~Torralba, ``Learning deep
  features for discriminative localization,'' in \emph{Proceedings of the IEEE
  Conference on Computer Vision and Pattern Recognition (CVPR)}, June 2016.

\bibitem{weak_SEAM}
Y.~Wang, J.~Zhang, M.~Kan, S.~Shan, and X.~Chen, ``Self-supervised equivariant
  attention mechanism for weakly supervised semantic segmentation,'' in
  \emph{Proceedings of the IEEE/CVF Conference on Computer Vision and Pattern
  Recognition (CVPR)}, June 2020.

\bibitem{scribble_ZhuangXH}
K.~Zhang and X.~Zhuang, ``Cyclemix: A holistic strategy for medical image
  segmentation from scribble supervision,'' in \emph{Proceedings of the
  IEEE/CVF Conference on Computer Vision and Pattern Recognition (CVPR)}, June
  2022, pp. 11\,656--11\,665.

\bibitem{scribble_transform}
X.~Liu, Q.~Yuan, Y.~Gao, K.~He, S.~Wang, X.~Tang, J.~Tang, and D.~Shen,
  ``Weakly supervised segmentation of covid19 infection with scribble
  annotation on ct images,'' \emph{Pattern Recognition}, vol. 122, p. 108341,
  2022.

\bibitem{bbox_BCM}
C.~Song, W.~Ouyang, and Z.~Zhang, ``Weakly supervised semantic segmentation via
  box-driven masking and filling rate shifting,'' \emph{IEEE Transactions on
  Pattern Analysis and Machine Intelligence}, vol.~45, no.~12, pp.
  15\,996--16\,012, 2023.

\bibitem{Cheng_2023_CVPR}
T.~Cheng, X.~Wang, S.~Chen, Q.~Zhang, and W.~Liu, ``Boxteacher: Exploring
  high-quality pseudo labels for weakly supervised instance segmentation,'' in
  \emph{Proceedings of the IEEE/CVF Conference on Computer Vision and Pattern
  Recognition (CVPR)}, June 2023, pp. 3145--3154.

\bibitem{Yu_2022_CVPR}
X.~Yu, P.~Chen, D.~Wu, N.~Hassan, G.~Li, J.~Yan, H.~Shi, Q.~Ye, and Z.~Han,
  ``Object localization under single coarse point supervision,'' in
  \emph{Proceedings of the IEEE/CVF Conference on Computer Vision and Pattern
  Recognition (CVPR)}, June 2022, pp. 4868--4877.

\bibitem{point_OD_point_teaching}
Y.~Ge, Q.~Zhou, X.~Wang, C.~Shen, Z.~Wang, and H.~Li, ``Point-teaching: Weakly
  semi-supervised object detection with point annotations,'' \emph{Proceedings
  of the AAAI Conference on Artificial Intelligence}, vol.~37, no.~1, pp.
  667--675, Jun. 2023.

\bibitem{point_Group_RCNN}
S.~Zhang, Z.~Yu, L.~Liu, X.~Wang, A.~Zhou, and K.~Chen, ``Group r-cnn for
  weakly semi-supervised object detection with points,'' in \emph{Proceedings
  of the IEEE/CVF Conference on Computer Vision and Pattern Recognition
  (CVPR)}, June 2022, pp. 9417--9426.

\bibitem{Wu_2024_CVPR}
W.~Wu, H.-S. Wong, S.~Wu, and T.~Zhang, ``Relational matching for weakly
  semi-supervised oriented object detection,'' in \emph{Proceedings of the
  IEEE/CVF Conference on Computer Vision and Pattern Recognition (CVPR)}, June
  2024, pp. 27\,800--27\,810.

\bibitem{LI2021107979}
X.~Li, H.~Ma, S.~Yi, Y.~Chen, and H.~Ma, ``Single annotated pixel based weakly
  supervised semantic segmentation under driving scenes,'' \emph{Pattern
  Recognition}, vol. 116, p. 107979, 2021.

\bibitem{7775087}
Y.~Wei, X.~Liang, Y.~Chen, X.~Shen, M.-M. Cheng, J.~Feng, Y.~Zhao, and S.~Yan,
  ``Stc: A simple to complex framework for weakly-supervised semantic
  segmentation,'' \emph{IEEE Transactions on Pattern Analysis and Machine
  Intelligence}, vol.~39, no.~11, pp. 2314--2320, 2017.

\bibitem{10597372}
Y.~Zhao, G.~Sun, Z.~Ling, A.~Zhang, and X.~Jia, ``Point-based weakly supervised
  deep learning for semantic segmentation of remote sensing images,''
  \emph{IEEE Transactions on Geoscience and Remote Sensing}, vol.~62, pp.
  1--16, 2024.

\bibitem{Fan2023}
J.~Fan and Z.~Zhang, ``Toward practical weakly supervised semantic segmentation
  via point-level supervision,'' \emph{International Journal of Computer
  Vision}, vol. 131, no.~12, pp. 3252--3271, Dec 2023.

\bibitem{semi_SEMI_survey}
J.~M. Duarte and L.~Berton, ``A review of semi-supervised learning for text
  classification,'' \emph{Artificial Intelligence Review}, pp. 1--69, 2023.

\bibitem{semi_GTA_TS}
Y.~Jin, J.~Wang, and D.~Lin, ``Semi-supervised semantic segmentation via gentle
  teaching assistant,'' in \emph{Advances in Neural Information Processing
  Systems}, vol.~35, 2022, pp. 2803--2816.

\bibitem{semi_double_tch_14}
Y.~Liu, Y.~Tian, Y.~Chen, F.~Liu, V.~Belagiannis, and G.~Carneiro, ``Perturbed
  and strict mean teachers for semi-supervised semantic segmentation,'' in
  \emph{Proceedings of the IEEE/CVF Conference on Computer Vision and Pattern
  Recognition (CVPR)}, June 2022, pp. 4258--4267.

\bibitem{semi_double_tch_21}
J.~Na, J.-W. Ha, H.~J. Chang, D.~Han, and W.~Hwang, ``Switching temporary
  teachers for semi-supervised semantic segmentation,'' \emph{Advances in
  Neural Information Processing Systems}, vol.~36, 2024.

\bibitem{semi_mixup_BCP}
Y.~Bai, D.~Chen, Q.~Li, W.~Shen, and Y.~Wang, ``Bidirectional copy-paste for
  semi-supervised medical image segmentation,'' in \emph{Proceedings of the
  IEEE/CVF Conference on Computer Vision and Pattern Recognition (CVPR)}, June
  2023, pp. 11\,514--11\,524.

\bibitem{semi_error_localization}
D.~Kwon and S.~Kwak, ``Semi-supervised semantic segmentation with error
  localization network,'' in \emph{Proceedings of the IEEE/CVF Conference on
  Computer Vision and Pattern Recognition (CVPR)}, June 2022, pp. 9957--9967.

\bibitem{semi_CANet}
C.~Zhao, S.~Xiang, Y.~Wang, Z.~Cai, J.~Shen, S.~Zhou, D.~Zhao, W.~Su, S.~Guo,
  and S.~Li, ``Context-aware network fusing transformer and v-net for
  semi-supervised segmentation of 3d left atrium,'' \emph{Expert Systems with
  Applications}, vol. 214, p. 119105, 2023.

\bibitem{semi_temporal}
S.~Laine and T.~Aila, ``Temporal ensembling for semi-supervised learning,''
  2017.

\bibitem{semi_pseudolabel}
D.~H. Lee, ``Pseudo-label : The simple and efficient semi-supervised learning
  method for deep neural networks,'' in \emph{Workshop on challenges in
  representation learning, ICML}, 2013.

\bibitem{semi_MCF}
Y.~Wang, B.~Xiao, X.~Bi, W.~Li, and X.~Gao, ``Mcf: Mutual correction framework
  for semi-supervised medical image segmentation,'' in \emph{Proceedings of the
  IEEE/CVF Conference on Computer Vision and Pattern Recognition (CVPR)}, June
  2023, pp. 15\,651--15\,660.

\bibitem{9941371}
X.~Yang, Z.~Song, I.~King, and Z.~Xu, ``A survey on deep semi-supervised
  learning,'' \emph{IEEE Transactions on Knowledge and Data Engineering},
  vol.~35, no.~9, pp. 8934--8954, 2023.

\bibitem{Liu_2022_CVPR}
C.~Liu, C.~Gao, F.~Liu, J.~Liu, D.~Meng, and X.~Gao, ``Ss3d:
  Sparsely-supervised 3d object detection from point cloud,'' in
  \emph{Proceedings of the IEEE/CVF Conference on Computer Vision and Pattern
  Recognition (CVPR)}, June 2022, pp. 8428--8437.

\bibitem{point_OD_point_DETR}
L.~Chen, T.~Yang, X.~Zhang, W.~Zhang, and J.~Sun, ``Points as queries: Weakly
  semi-supervised object detection by points,'' in \emph{Proceedings of the
  IEEE/CVF Conference on Computer Vision and Pattern Recognition (CVPR)}, June
  2021, pp. 8823--8832.

\bibitem{point_OD_point_DETR_3DOD}
D.~Zhang, D.~Liang, Z.~Zou, J.~Li, X.~Ye, Z.~Liu, X.~Tan, and X.~Bai, ``A
  simple vision transformer for weakly semi-supervised 3d object detection,''
  in \emph{Proceedings of the IEEE/CVF International Conference on Computer
  Vision (ICCV)}, October 2023, pp. 8373--8383.

\bibitem{point_WangHong_MICCAI}
H.~Ji, H.~Liu, Y.~Li, J.~Xie, N.~He, Y.~Huang, D.~Wei, X.~Chen, L.~Shen, and
  Y.~Zheng, ``Point beyond class: A benchmark for weakly semi-supervised
  abnormality localization in chest x-rays,'' in \emph{Medical Image Computing
  and Computer Assisted Intervention}.\hskip 1em plus 0.5em minus 0.4em\relax
  Cham: Springer Nature Switzerland, 2022, pp. 249--260.

\bibitem{point_WangHong_MIA}
Y.~Shi, H.~Wang, H.~Ji, H.~Liu, Y.~Li, N.~He, D.~Wei, Y.~Huang, Q.~Dai, J.~Wu,
  X.~Chen, Y.~Zheng, and H.~Yu, ``A deep weakly semi-supervised framework for
  endoscopic lesion segmentation,'' \emph{Medical Image Analysis}, vol.~90, p.
  102973, 2023.

\bibitem{point_SAM}
A.~Kirillov, E.~Mintun, N.~Ravi, H.~Mao, C.~Rolland, L.~Gustafson, T.~Xiao,
  S.~Whitehead, A.~C. Berg, W.-Y. Lo, P.~Dollar, and R.~Girshick, ``Segment
  anything,'' pp. 4015--4026, October 2023.

\bibitem{PointSupervision}
A.~Bearman, O.~Russakovsky, V.~Ferrari, and L.~Fei-Fei, ``What's the point:
  Semantic segmentation with point supervision,'' in \emph{Computer Vision --
  ECCV 2016}, B.~Leibe, J.~Matas, N.~Sebe, and M.~Welling, Eds.\hskip 1em plus
  0.5em minus 0.4em\relax Cham: Springer International Publishing, 2016, pp.
  549--565.

\bibitem{Cheng_2022_CVPR}
B.~Cheng, O.~Parkhi, and A.~Kirillov, ``Pointly-supervised instance
  segmentation,'' in \emph{Proceedings of the IEEE/CVF Conference on Computer
  Vision and Pattern Recognition (CVPR)}, June 2022, pp. 2617--2626.

\bibitem{other_mixup}
H.~Zhang, M.~Cisse, Y.~N. Dauphin, and D.~Lopez-Paz, ``mixup: Beyond empirical
  risk minimization,'' 2018.

\bibitem{weak_ADELE}
S.~Liu, K.~Liu, W.~Zhu, Y.~Shen, and C.~Fernandez-Granda, ``Adaptive
  early-learning correction for segmentation from noisy annotations,'' in
  \emph{Proceedings of the IEEE/CVF Conference on Computer Vision and Pattern
  Recognition (CVPR)}, June 2022, pp. 2606--2616.

\bibitem{Chen_2022_CVPR}
Z.~Chen, T.~Wang, X.~Wu, X.-S. Hua, H.~Zhang, and Q.~Sun, ``Class re-activation
  maps for weakly-supervised semantic segmentation,'' in \emph{Proceedings of
  the IEEE/CVF Conference on Computer Vision and Pattern Recognition (CVPR)},
  June 2022, pp. 969--978.

\bibitem{Chen2_2022_CVPR}
Q.~Chen, L.~Yang, J.-H. Lai, and X.~Xie, ``Self-supervised image-specific
  prototype exploration for weakly supervised semantic segmentation,'' in
  \emph{Proceedings of the IEEE/CVF Conference on Computer Vision and Pattern
  Recognition (CVPR)}, June 2022, pp. 4288--4298.

\bibitem{wang2019boundary}
B.~Wang, G.~Qi, S.~Tang, T.~Zhang, Y.~Wei, L.~Li, and Y.~Zhang, ``Boundary
  perception guidance: A scribble-supervised semantic segmentation approach,''
  in \emph{IJCAI International joint conference on artificial intelligence},
  2019.

\bibitem{scribble_RGB_D1}
Y.~Xu, X.~Yu, J.~Zhang, L.~Zhu, and D.~Wang, ``Weakly supervised rgb-d salient
  object detection with prediction consistency training and active scribble
  boosting,'' \emph{IEEE Transactions on Image Processing}, vol.~31, pp.
  2148--2161, 2022.

\bibitem{point_supervision_infrared}
X.~Ying, L.~Liu, Y.~Wang, R.~Li, N.~Chen, Z.~Lin, W.~Sheng, and S.~Zhou,
  ``Mapping degeneration meets label evolution: Learning infrared small target
  detection with single point supervision,'' in \emph{Proceedings of the
  IEEE/CVF Conference on Computer Vision and Pattern Recognition (CVPR)}, June
  2023, pp. 15\,528--15\,538.

\bibitem{point_supervision_SOD_floodfill}
S.~Gao, W.~Zhang, Y.~Wang, Q.~Guo, C.~Zhang, Y.~He, and W.~Zhang,
  ``Weakly-supervised salient object detection using point supervision,'' in
  \emph{AAAI}, 2022.

\bibitem{point_supervision_SOD_floodfill_2}
S.~Gao, H.~Xing, W.~Zhang, Y.~Wang, Q.~Guo, and W.~Zhang, ``Weakly supervised
  video salient object detection via point supervision,'' in \emph{Proceedings
  of the 30th ACM International Conference on Multimedia}, 2022, p.
  3656–3665.

\bibitem{semi_EMA}
A.~Tarvainen and H.~Valpola, ``Mean teachers are better role models:
  Weight-averaged consistency targets improve semi-supervised deep learning
  results,'' \emph{Advances in Neural Information Processing Systems}, vol.~30,
  2017.

\bibitem{10120949}
M.~Lee, S.~Lee, J.~Lee, and H.~Shim, ``Saliency as pseudo-pixel supervision for
  weakly and semi-supervised semantic segmentation,'' \emph{IEEE Transactions
  on Pattern Analysis and Machine Intelligence}, vol.~45, no.~10, pp.
  12\,341--12\,357, 2023.

\bibitem{point_supervision_devil}
B.~Kim, J.~Jeong, D.~Han, and S.~J. Hwang, ``The devil is in the points: Weakly
  semi-supervised instance segmentation via point-guided mask representation,''
  in \emph{Proceedings of the IEEE/CVF Conference on Computer Vision and
  Pattern Recognition (CVPR)}, June 2023, pp. 11\,360--11\,370.

\bibitem{backbone_DeepLabV3Plus}
L.-C. Chen, Y.~Zhu, G.~Papandreou, F.~Schroff, and H.~Adam, ``Encoder-decoder
  with atrous separable convolution for semantic image segmentation,'' in
  \emph{Proceedings of the European Conference on Computer Vision (ECCV)},
  September 2018.

\bibitem{backbone_PSPNet}
H.~Zhao, J.~Shi, X.~Qi, X.~Wang, and J.~Jia, ``Pyramid scene parsing network,''
  in \emph{Proceedings of the IEEE Conference on Computer Vision and Pattern
  Recognition (CVPR)}, July 2017.

\bibitem{backbone_Segnet}
V.~Badrinarayanan, A.~Kendall, and R.~Cipolla, ``Segnet: A deep convolutional
  encoder-decoder architecture for image segmentation,'' \emph{IEEE
  Transactions on Pattern Analysis and Machine Intelligence}, vol.~39, no.~12,
  pp. 2481--2495, 2017.

\bibitem{backbone_segformer}
E.~Xie, W.~Wang, Z.~Yu, A.~Anandkumar, J.~M. Alvarez, and P.~Luo, ``Segformer:
  Simple and efficient design for semantic segmentation with transformers,'' in
  \emph{Advances in Neural Information Processing Systems}, M.~Ranzato,
  A.~Beygelzimer, Y.~Dauphin, P.~Liang, and J.~W. Vaughan, Eds., vol.~34.\hskip
  1em plus 0.5em minus 0.4em\relax Curran Associates, Inc., 2021, pp.
  12\,077--12\,090.

\bibitem{Laradji_2018_ECCV}
I.~H. Laradji, N.~Rostamzadeh, P.~O. Pinheiro, D.~Vazquez, and M.~Schmidt,
  ``Where are the blobs: Counting by localization with point supervision,'' in
  \emph{Proceedings of the European Conference on Computer Vision (ECCV)},
  September 2018.

\bibitem{pmlrverma19a}
V.~Verma, A.~Lamb, C.~Beckham, A.~Najafi, I.~Mitliagkas, D.~Lopez-Paz, and
  Y.~Bengio, ``Manifold mixup: Better representations by interpolating hidden
  states,'' in \emph{Proceedings of the 36th International Conference on
  Machine Learning}, ser. Proceedings of Machine Learning Research,
  K.~Chaudhuri and R.~Salakhutdinov, Eds., vol.~97.\hskip 1em plus 0.5em minus
  0.4em\relax PMLR, 09--15 Jun 2019, pp. 6438--6447.

\bibitem{Guo_Mao_Zhang_2019}
H.~Guo, Y.~Mao, and R.~Zhang, ``Mixup as locally linear out-of-manifold
  regularization,'' \emph{Proceedings of the AAAI Conference on Artificial
  Intelligence}, vol.~33, no.~01, pp. 3714--3722, Jul. 2019.

\bibitem{10378660}
Z.~Liu, S.~Zheng, X.~Sun, Z.~Zhu, Y.~Zhao, X.~Yang, and Y.~Zhao, ``The devil is
  in the boundary: Boundary-enhanced polyp segmentation,'' \emph{IEEE
  Transactions on Circuits and Systems for Video Technology}, vol.~34, no.~7,
  pp. 5414--5423, 2024.

\bibitem{Chen_2021_CVPR}
X.~Chen, Y.~Yuan, G.~Zeng, and J.~Wang, ``Semi-supervised semantic segmentation
  with cross pseudo supervision,'' in \emph{Proceedings of the IEEE/CVF
  Conference on Computer Vision and Pattern Recognition (CVPR)}, June 2021, pp.
  2613--2622.

\bibitem{bmvc_DASD}
G.~French, S.~Laine, T.~Aila, M.~Mackiewicz, and G.~Finlayson,
  ``Semi-supervised semantic segmentation needs strong, varied perturbations,''
  in \emph{British Machine Vision Conference (BMVC)}, Sep 2020.

\bibitem{Kvasir_SEG}
D.~Jha, P.~H. Smedsrud, M.~A. Riegler, P.~Halvorsen, T.~De~Lange, D.~Johansen,
  and H.~D. Johansen, ``Kvasir-seg: A segmented polyp dataset,'' in
  \emph{MultiMedia modeling: 26th international conference, MMM 2020, Daejeon,
  South Korea, January 5--8, 2020, proceedings, part II 26}.\hskip 1em plus
  0.5em minus 0.4em\relax Springer, 2020, pp. 451--462.

\bibitem{CVC_ClinicDB_1}
D.~Vázquez, J.~Bernal, F.~J. Sánchez, G.~Fernández-Esparrach, A.~M. López,
  A.~Romero, M.~Drozdzal, and A.~Courville, ``A benchmark for endoluminal scene
  segmentation of colonoscopy images,'' \emph{Journal of Healthcare
  Engineering}, vol. 2017, no.~1, p. 4037190, 2017.

\bibitem{CVC_ClinicDB_2}
J.~Bernal, F.~J. Sánchez, G.~Fernández-Esparrach, D.~Gil, C.~Rodríguez, and
  F.~Vilariño, ``Wm-dova maps for accurate polyp highlighting in colonoscopy:
  Validation vs. saliency maps from physicians,'' \emph{Computerized Medical
  Imaging and Graphics}, vol.~43, pp. 99--111, 2015.

\bibitem{ETIS_Larib}
J.~Silva, A.~Histace, O.~Romain, X.~Dray, and B.~Granado, ``Toward embedded
  detection of polyps in wce images for early diagnosis of colorectal cancer,''
  \emph{International journal of computer assisted radiology and surgery},
  vol.~9, pp. 283--293, 2014.

\bibitem{sanderson2022fcn}
E.~Sanderson and B.~J. Matuszewski, ``Fcn-transformer feature fusion for polyp
  segmentation,'' in \emph{Annual conference on medical image understanding and
  analysis}.\hskip 1em plus 0.5em minus 0.4em\relax Springer, 2022, pp.
  892--907.

\bibitem{nguyen2021ccbanet}
T.-C. Nguyen, T.-P. Nguyen, G.-H. Diep, A.-H. Tran-Dinh, T.~V. Nguyen, and
  M.-T. Tran, ``Ccbanet: cascading context and balancing attention for polyp
  segmentation,'' in \emph{Medical Image Computing and Computer Assisted
  Intervention--MICCAI 2021: 24th International Conference, Strasbourg, France,
  September 27--October 1, 2021, Proceedings, Part I 24}.\hskip 1em plus 0.5em
  minus 0.4em\relax Springer, 2021, pp. 633--643.

\bibitem{Zhou_2022_CVPR}
K.~Zhou, J.~Yang, C.~C. Loy, and Z.~Liu, ``Conditional prompt learning for
  vision-language models,'' in \emph{Proceedings of the IEEE/CVF Conference on
  Computer Vision and Pattern Recognition (CVPR)}, June 2022, pp.
  16\,816--16\,825.

\bibitem{10.1145/3560815}
P.~Liu, W.~Yuan, J.~Fu, Z.~Jiang, H.~Hayashi, and G.~Neubig, ``Pre-train,
  prompt, and predict: A systematic survey of prompting methods in natural
  language processing,'' \emph{ACM Computing Surveys}, vol.~55, no.~9, 2023.

\bibitem{SAM_not_perfect}
W.~Ji, J.~Li, Q.~Bi, T.~Liu, W.~Li, and L.~Cheng, ``Segment anything is not
  always perfect: An investigation of sam on different real-world
  applications,'' \emph{Machine Intelligence Research}, vol.~21, no.~4, pp.
  617--630, Aug 2024.

\end{thebibliography}

\end{document}